\documentclass{article}

\usepackage{microtype}
\usepackage{graphicx}
\usepackage{subcaption}
\usepackage{booktabs} 
\usepackage{xcolor} 
\usepackage{tcolorbox}
\usepackage{graphicx}
\usepackage{tcolorbox}
\tcbuselibrary{skins,breakable}
\usepackage{booktabs}
\usepackage{colortbl}
\usepackage{xcolor}
\usepackage{graphicx}
\usepackage{subcaption}
\usepackage{multirow}
\usepackage{hyperref}

\usepackage[accepted]{icml2026}

\usepackage{amsmath}
\usepackage{amssymb}
\usepackage{mathtools}
\usepackage{amsthm}

\usepackage[capitalize,noabbrev]{cleveref}

\theoremstyle{plain}

\theoremstyle{definition}

\theoremstyle{remark}

\usepackage[textsize=tiny]{todonotes}

\icmltitlerunning{Uncertainty-Aware Clarification in LLM Agents with Information Gain}

\begin{document}

\twocolumn[
  \icmltitle{Uncertainty-Aware Clarification in LLM Agents with Information Gain}

  \icmlsetsymbol{equal}{*}

  \begin{icmlauthorlist}

    \icmlauthor{Mengyi Deng}{hkustgz}
    \icmlauthor{Zhiwei Li}{hkustgz}
    \icmlauthor{Xin Li}{hkustgz}
    \icmlauthor{Tingyu Zhu}{hkustgz}
    \icmlauthor{Ying Zhao}{jlu}
    \icmlauthor{Zhijiang Guo}{hkustgz,hkust}
    \icmlauthor{Wei Wang}{hkustgz,hkust}

  \end{icmlauthorlist}

  \icmlaffiliation{hkustgz}{
  Information Hub,
  HKUST(GZ)
  }

  \icmlaffiliation{hkust}{
  HKUST
  }

  \icmlaffiliation{jlu}{
  Jilin University
  }

\icmlcorrespondingauthor{Wei Wang}{weiwcs@ust.hk}
\icmlcorrespondingauthor{Zhijiang Guo}{zhijiangguo@hkust-gz.edu.cn}
\icmlcorrespondingauthor{Mengyi Deng}{mdeng974@connect.hkust-gz.edu.cn}

  \icmlkeywords{Machine Learning, ICML}

  \vskip 0.3in
]



\printAffiliationsAndNotice{}  

\begin{abstract}

Large Language Model (LLM) agents often operate under underspecified user instructions, where latent uncertainty over user intent leads to erroneous tool actions. To address this challenge, we propose a goal-oriented clarification framework that aligns clarification behavior with ambiguity resolution. Central to our approach is the Information Gain Reward, a metric that quantifies the utility of clarification questions by measuring the Bayesian belief update towards the ground-truth goal induced by the clarification exchange. We train the clarifier (LLM) using this reward to optimize for high information gain, ensuring that clarifications effectively reduce uncertainty and improve task completion within the agent-tool-user environment. We validate our framework within a clarification-enhanced $\tau$-Bench environment, conducting cross-agent evaluations across five heterogeneous backbones. Empirical results demonstrate that our method consistently improves the success rate by 3.7\% over the no-clarification baseline, while adding only 0.3 total interaction steps on average.

\end{abstract}

\noindent\textbf{Code:} \href{https://github.com/Demi-deng2/IG-clarifier}{github.com/Demi-deng2/IG-clarifier}

\section{Introduction}
Large Language Model (LLM) agents have recently shown significant proficiency in utilizing external tools and performing multi-step decision-making across a diverse set of real-world applications~\cite{qu2025tool,yao2022react,wolflein2025llm}. Through their interaction with various external tools, these agents are capable of completing complex tasks that involve planning, reasoning, and iterative feedback~\cite{wolflein2025llm}. In practical interactive contexts, however, user instructions are often \emph{underspecified} or vague, leaving essential aspects of the user’s true intent implicit~\cite{yehudai2025survey,qi2025agentif}. Such ambiguous user demands present a fundamental challenge for tool-using agents: premature or inaccurate tool actions may lead to irreversible mistakes, fragile execution paths, and task failure~\cite{wang2024gta}. 

\begin{figure}
    \centering
    \includegraphics[width=1\linewidth]{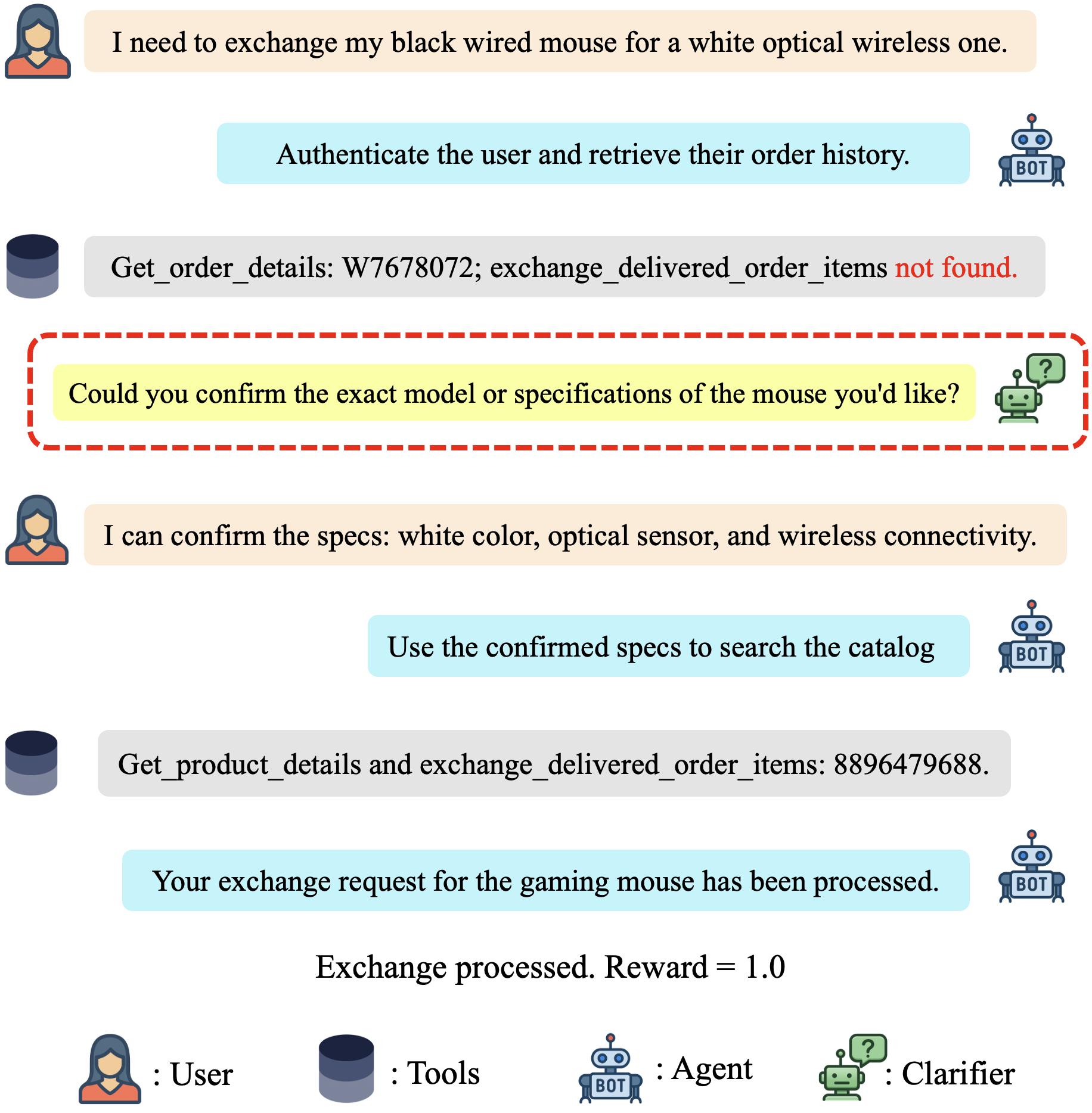}
    \caption{An example of clarification in a $\tau$-retail trajectory. When an initial tool call fails due to missing or underspecified information, the Clarifier poses a targeted follow-up question (highlighted) to elicit the required constraints from the user. The additional information provided through this exchange allows the agent to proceed with a corrected tool invocation and complete the task.}
    \label{fig:clarify}
\end{figure}

A common approach to addressing underspecified user instructions is issuing clarification questions, as shown in Figure~\ref{fig:clarify}, to resolve ambiguity before action~\cite{clark1996using, zou2023users}. While human assistants naturally identify missing information and seek clarification, this ability remains underdeveloped in LLM-based agents~\cite{rao2018learning}. A key challenge is that LLMs are primarily trained to answer questions rather than to determine when clarification is necessary or how to formulate it. Consequently, existing training pipelines provide scant guidance on the necessity or effectiveness of clarification~\cite{fu2020survey, ouyang2022training}, often leaving such behaviors implicit or entangled with internal reasoning~\cite{suri2025structured}. Moreover, the difficulty in training LLMs for clarification stems from a misalignment in supervision: existing annotations typically reward surface-level fluency while failing to capture a question’s capacity to reduce requirement uncertainty~\cite{zhang2023instruction}. Without an explicit signal for uncertainty reduction, models default to favoring generic, safe questions that provide limited informational value~\cite{szymanski2025limitations, son2024llm}, struggling to evolve toward truly targeted information-seeking behaviors.

To address these challenges and move beyond subjective supervision, we reframe the clarification problem within the context of uncertainty-aware learning. Motivated by this perspective, we propose an information-theoretic framework that quantifies the utility of a clarification question by measuring its impact on reducing uncertainty about the user's true goal. Specifically, we model the clarification process as a Bayesian belief update and quantify the utility of a question by measuring the shift in the model's probability mass towards the ground-truth goal following the clarification exchange. This formulation provides a performance-oriented objective that directly links questioning strategies to effective ambiguity resolution. We then utilize this intrinsic reward signal to drive Decoupled Advantage Policy Optimization (DAPO)~\cite{yu2025dapo}, guiding the agent to evolve questioning strategies that are explicitly optimized for missing information recovery rather than surface-level conversational behaviors.

This paper studies clarification as an execution-grounded information acquisition problem, where the agent intervenes only when asking a question is expected to reduce uncertainty about the latent user goal and improve downstream tool use. We evaluate this formulation in a Clarifier-augmented $\tau$-Bench setting, where user intent is partially observable and tool feedback exposes execution failures, requiring the agent to recover through interactive clarification rather than direct access to the full goal. We further validate the framework through comprehensive analyzes of training dynamics, the upper bounds of agent and clarification effectiveness, trigger frequency, and cross-agent generalization.

\section{Related Work}
\paragraph{Tool-Augmented Agents.}
Large language models have increasingly been framed as tool-using agents capable of performing multi-step reasoning and decision-making. These agents utilize various processes, such as planning~\cite{yao2022react, schick2023toolformer}, tool invocation~\cite{yuan2025easytool, zhu2025agentar, wu2024avatar}, and feedback-driven execution~\cite{lu2025toolsandbox, wu2025agentic, liu2023dynamic}. To evaluate agent behavior in complex, tool-rich settings, various benchmarks and environments have been proposed. For example, \cite{qu2025tool} focuses on tool efficiency, \cite{liu2023agentbench} provides a broad framework for agent task evaluation, and \cite{andriushchenko2024agentharm} addresses robustness in adversarial contexts. These diverse benchmarks~\cite{wang2024gta,andriushchenko2024agentharm,xu2024theagentcompany,li2024embodied,wu2024streambench} offer valuable insights into agent performance across different application domains. However, while existing agent frameworks primarily optimize tool invocation for well-defined tasks, they lack explicit mechanisms to handle ambiguity in real-world instructions, often entangling clarification with execution. To address this, our work leverages the dynamic interaction capabilities of $\tau$-Bench~\cite{yao2024tau} to explicitly optimize clarification strategies.

\paragraph{Clarification Learning and Alignment Signals.}
Clarification questions serve as a critical mechanism for resolving ambiguity. While foundational studies emphasized their necessity for grounding in human communication~\cite{clark1996using,traum2003information,yizhou2024clarinet,zhang2024ask}, recent advances have extended this mechanism to conversational search~\cite{aliannejadi2019asking} and active uncertainty reduction strategies~\cite{zou2023users,xia2025selection,li2024active}. While current efforts primarily focus on evaluation~\cite{zhou2025passive} or static question selection~\cite{choudhury2025bed}, they often overlook the learning of clarification policies for tool-using agents. In parallel, IGPO~\cite{wang2025information} formulates information-gain-based optimization for multi-turn agents under the assumption that the target answer is known in advance. Compared with existing methods operate with predefined answer targets or fixed clarification candidates, our setting considers tool-using agents that infer latent user goals online, condition on tool state and execution feedback, handle free-form user responses, and learn an amortized clarification policy rather than only selecting from static questions or optimizing against a known answer slot.

Although emerging policy optimization frameworks like GRPO~\cite{shao2024deepseekmath,wang2025survey} and DAPO~\cite{yu2025dapo} offer a robust mechanism to learn such behaviors, applying them to clarification is challenging: standard alignment approaches typically rely on human or LLM-based judges that prioritize surface-level text quality over informational value~\cite{chaudhari2025rlhf}. To address this, we propose an information-theoretic framework that explicitly measures a question's ability to resolve underspecified user intent. Specifically, we optimize for Expected Information Gain (EIG) to quantify goal uncertainty reduction. This enables policy optimization to explicitly target uncertainty reduction, maximizing information gain to facilitate successful task execution.
\begin{figure*}
    \centering
    \includegraphics[width=1\linewidth]{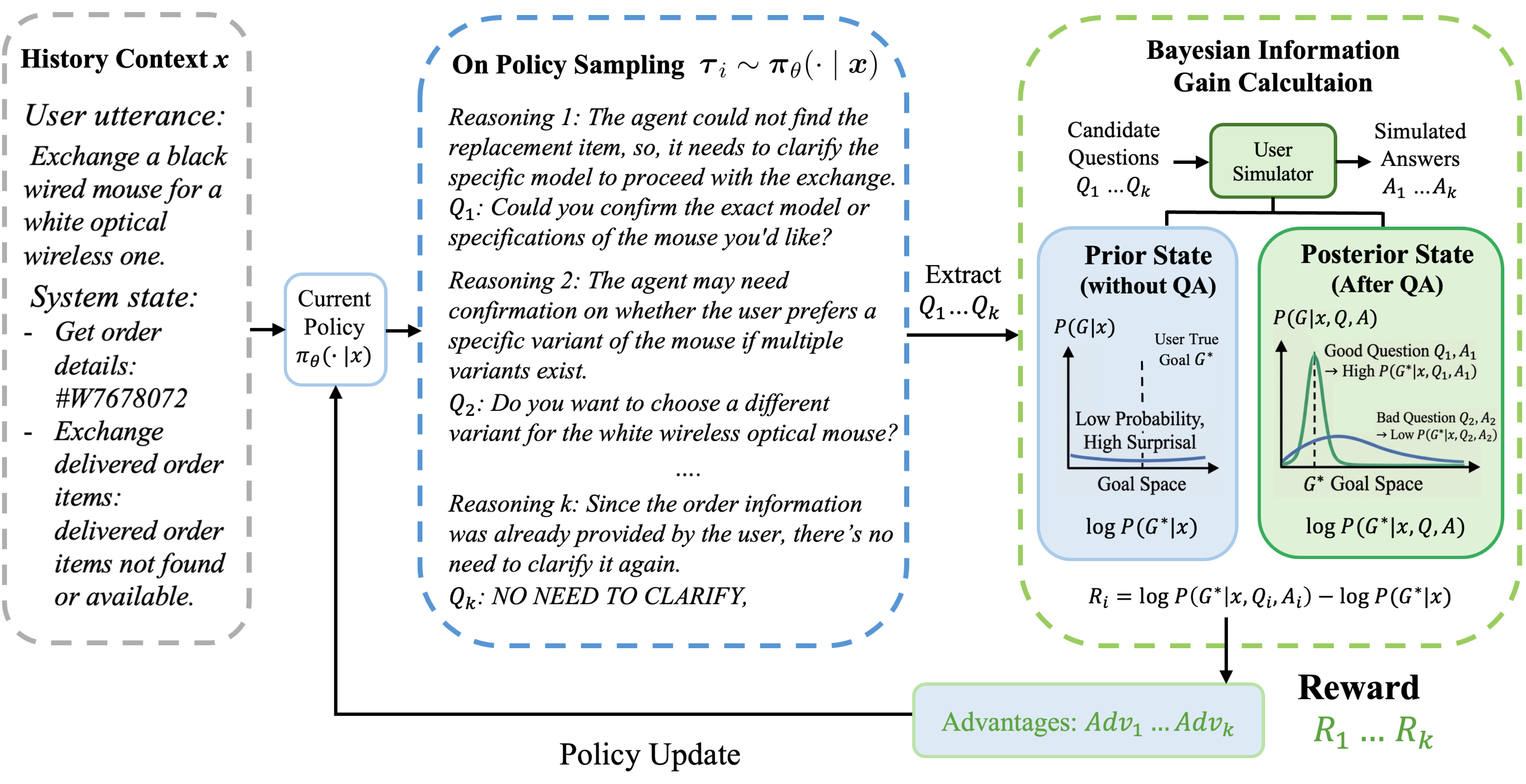}
    \caption{Overview of the Amortized Bayesian Experimental Design framework. The model performs on-policy sampling to generate candidate questions. These candidates are evaluated by our Belief Update Reward, which quantifies efficacy as the shift in the teacher-forced log-likelihood of the ground-truth goal $G^*$ (Bayesian Belief Update). Acting as an amortized surrogate for the intractable expected information gain, this signal guides the DAPO update, steering the policy towards clarifications that maximally reduce goal uncertainty.}
    \label{framework2}
\end{figure*}

\section{Method}
\label{method}
In this section, we present a framework for learning alignment-driven clarification policies. We first introduce the execution-grounded interaction protocol within $\tau$-Bench (Section~\ref{sec:env}), followed by the data construction process (Section~\ref{sec:data}). Section~\ref{sec:reward} details the Amortized Bayesian Experimental Design and its role in formulating the Information Gain Reward, while Section~\ref{alignment} explains how the DAPO algorithm uses this reward to optimize the clarifier for resolving task ambiguity.

\subsection{Environment and Interaction Protocol}
\label{sec:env}
We conduct our study on $\tau$-Bench~\cite{yao2024tau}, which provides a partially observable Markov decision process (POMDP) based environment for tool-agent interaction.
To handle user ambiguity, we extend this interaction loop with a pluggable and LLM-based clarifier module, as illustrated in Figure~\ref{fig:clarify}. When the agent issues a tool call $a_t \in \mathcal{A}_{\text{tool}}$ and receives the corresponding observation $o_t$, the clarifier assesses whether the execution feedback reveals latent ambiguity or missing task-critical information. If so, it generates a concise follow-up question; otherwise, no clarification turn is produced. We focus on learning a clarification policy that optimizes the decision of whether to intervene and what information to solicit based on execution feedback. This process yields an interaction trajectory defined as: 
\[
\tau = \Big(U,\; (a_t,\; o_t,\; Q_t,\; A_t)_{t=1}^{T},\; G\Big),
\]
where $U$ denotes the initial fuzzy instruction and $G$ represents the user's latent ground-truth goal. While $G$ remains unobserved by the agent during interaction, it guides the user simulator's responses and serves as the reference for computing our information-gain rewards. At each step $t$, the agent executes a tool action $a_t$ and observes $o_t$. Conditioned on this execution feedback, the clarifier may issue a clarification question $Q_t$ and receive a corresponding response $A_t$ from the user simulator before the next action is taken. If no clarification is made at step $t$, we set $Q_t = A_t = \varnothing$. These trajectories serve as the foundation for optimizing our clarifier policy. 

\subsection{Data Construction}
\label{sec:data}
To curate the dataset for policy optimization, we process interaction trajectories from $\tau$-Bench, restricting our scope to the first 16 agent steps to focus on ambiguity resolution. From this filtered subset, we retain a final dataset of $2{,}676$ step-level instances, where each data point consists of the dialogue history $x$ and a task-defined user goal $G$ provided by $\tau$-Bench. We further normalize the raw user goals to obtain a clean, canonical goal representation $G^*$, which serves as the supervision target for reward computation throughout this paper. Details of the data construction process and the goal transfer from $G$ to $G^*$ are provided in Appendix~\ref{app:data-construct}.

\subsection{Amortized Bayesian Experimental Design}
\label{sec:reward}

To learn \emph{when} and \emph{what} to clarify, we formulate clarification generation from a Bayesian Experimental Design (BED)~\cite{chaloner1995bayesian} perspective and train a clarifier to propose questions that are expected to reduce uncertainty about the user's latent goal $G^*$. Rather than directly optimizing this objective, which is intractable in interactive settings, we approximate the experimental design principle through an amortized training pipeline.

\paragraph{Theoretical Objective.}
The amortized training pipeline described above can be viewed as an approximation to the following idealized Bayesian objective.  Formally, let $G^*$ be the ground-truth goal, $x$ be the interaction context, and $P(A \mid Q,x)$ denote the answer distribution induced by the user simulator after question $Q$ is asked in context $x$. An optimal clarification $Q^*$ maximizes the \textit{Expected Information Gain (EIG)}:
\begin{footnotesize}
\begin{equation}
    I(G^*; A \mid Q, x) = \mathbb{E}_{A \sim P(\cdot|Q, x)} \Big[ H(G^* \mid x) - H(G^* \mid x, Q, A) \Big]
\end{equation}
\end{footnotesize}
where $H(\cdot)$ denotes entropy. Direct maximization of this objective is computationally intractable. Specifically, calculating the EIG requires integrating over the vast, open-ended space of latent user goals and potential responses to estimate the posterior entropy. Such high-dimensional inference is prohibitively expensive for every clarification candidate.
\paragraph{Amortized Optimization via Belief Update.}
To address this, we propose an \textbf{amortized} approach. We optimize a question policy $\pi_\theta(Q \mid x)$ to internalize the experimental design process and directly generate high-information questions. Separately, we use the same model in teacher-forcing mode as a belief scorer $P_\theta(G^* \mid \cdot)$, with gradients detached during reward computation. We define the reward $R_t$ as a \textit{pointwise} information-gain proxy. To ensure the reward is invariant to goal complexity, we formulate it as the shift in the length-normalized teacher-forced log-likelihood of $G^*$:
\begin{equation}
    \begin{split}
        R_t(x_t, Q_t, A_t) &= \underbrace{\frac{1}{L} \sum_{j=1}^{L} \log P_{\theta}(g^*_j \mid x_t, Q_t, A_t, g^*_{<j})}_{\text{Posterior Belief}} \\
        &\quad - \underbrace{\frac{1}{L} \sum_{j=1}^{L} \log P_{\theta}(g^*_j \mid x_t, g^*_{<j})}_{\text{Prior Belief}}.
    \end{split}
    \label{eq:belief_update}
\end{equation}
Here, $G^* = (g^*_1, \dots, g^*_L)$ represents the token sequence of the ground-truth goal. The term $\log P_\theta(\cdot)$ denotes the probability of the next token computed via teacher forcing. By normalizing these log-probabilities (dividing by length $L$), $R_t$ quantifies the Bayesian Belief Update: a positive value indicates that the clarification exchange $(Q_t, A_t)$ has effectively concentrated the model's probability mass on the true goal sequence. To estimate this belief update in practice, any sufficiently capable LLM could be used to estimate these likelihood terms. In this work,  we use the same backbone as the clarifier policy to maintain consistency between belief representation and policy behavior. Since the policy is trained to reduce its own uncertainty over the user goal, evaluating likelihood shifts within the same model ensures that the reward reflects the policy’s internal belief update.

Mathematically, as derived in Appendix~\ref{app:proof}, this reward corresponds to the Pointwise Mutual Information (PMI) between the response and the ground truth. It can be expressed as the log-importance weight $\log \frac{P(A \mid G^*, Q, x)}{P(A \mid Q, x)}$, which effectively quantifies how much the observed answer $A$ precisely distinguishes the true goal $G^*$ from the general hypothesis space. Crucially, when the clarifier decides not to intervene (i.e., \texttt{No need to ask} token), the posterior belief remains identical to the prior, naturally yielding a reward of zero ($R_t = 0$).

\paragraph{Strict User Simulator for Training.}
The validity of the belief update relies heavily on the quality of the simulated response $A_t$. If the user simulator is too helpful (revealing the goal even for generic questions), the clarifier tends to hack the reward without learning valid clarification strategies.
Therefore, we employ a Strict User Simulator during training. Unlike the relaxed user prompt used in $\tau$-Bench, our strict simulator reveals task-specific information \textit{only} when the question $Q_t$ is specific and relevant. This ensures that high rewards are assigned exclusively to questions that genuinely necessitate clarification. A detailed comparison of the simulator prompts is provided in Appendix~\ref{app:user_prompts}.

\subsection{Optimization via DAPO}
\label{alignment}
As illustrated in Figure~\ref{framework2}, the clarifier operates on the current interaction context $x$, which consists of the accumulated dialogue and execution history and serves as the input to the clarifier LLM. Conditioned on this context, the policy performs on-policy sampling to generate a set of candidate clarification questions $\{Q_1,\dots,Q_K\}$.
Each candidate is answered by the Strict User Simulator, yielding responses $\{A_1,\dots,A_K\}$. We evaluate the resulting trajectories using the Belief Update Reward $R_t$ derived in Eq.~\ref{eq:belief_update}. Finally, these rewards are converted into group-relative advantages to update the policy via DAPO~\cite{yu2025dapo}, maximizing the expected cumulative information gain.
\begin{figure*}
    \centering
    \begin{subfigure}{0.492\linewidth}
        \centering
        \includegraphics[width=\linewidth]{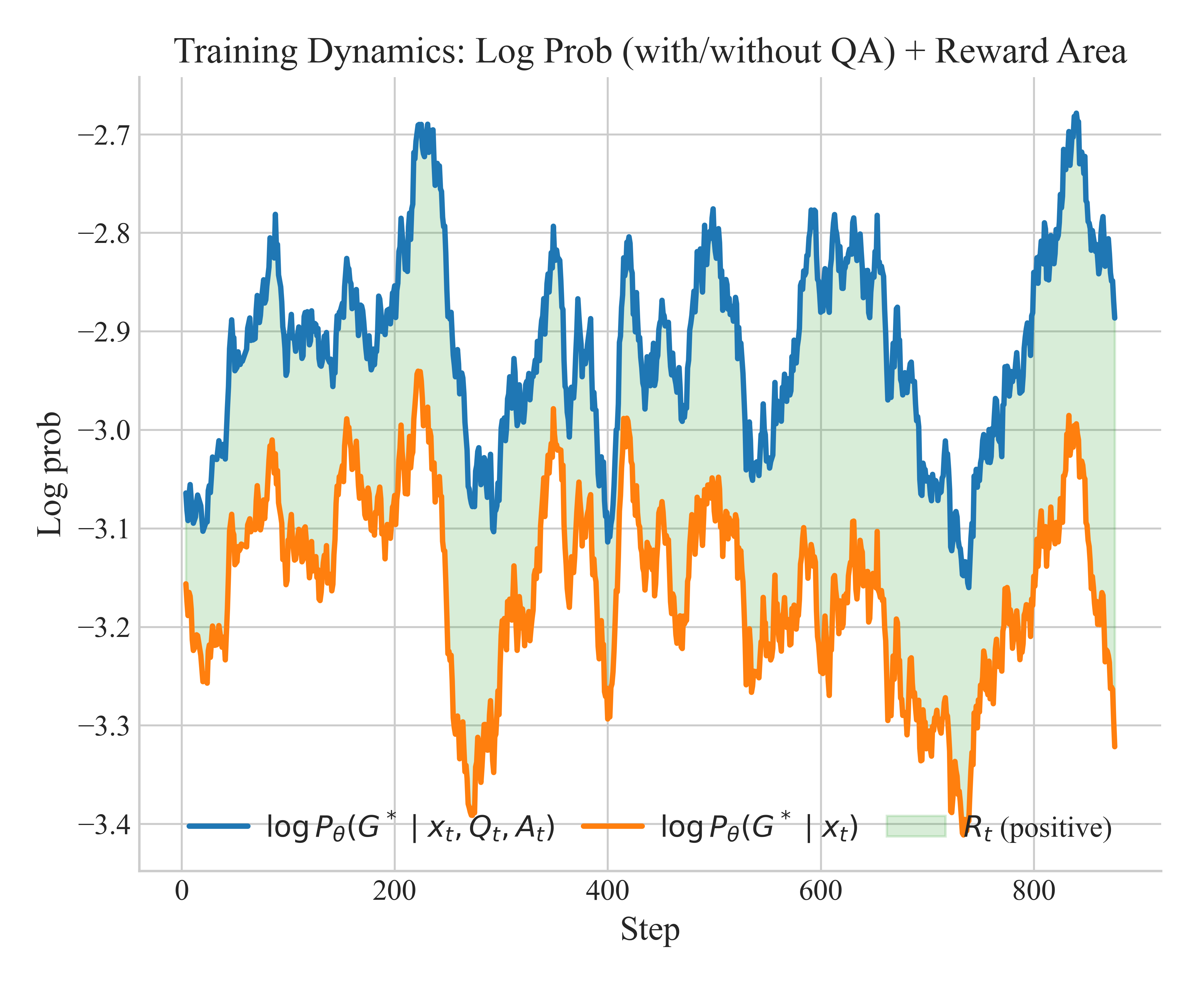}
        \caption{Log probabilities with and without clarification; the shaded difference denotes the information gain reward.}
    \end{subfigure}
    \hfill
    \begin{subfigure}{0.492\linewidth}
        \centering
        \includegraphics[width=\linewidth]{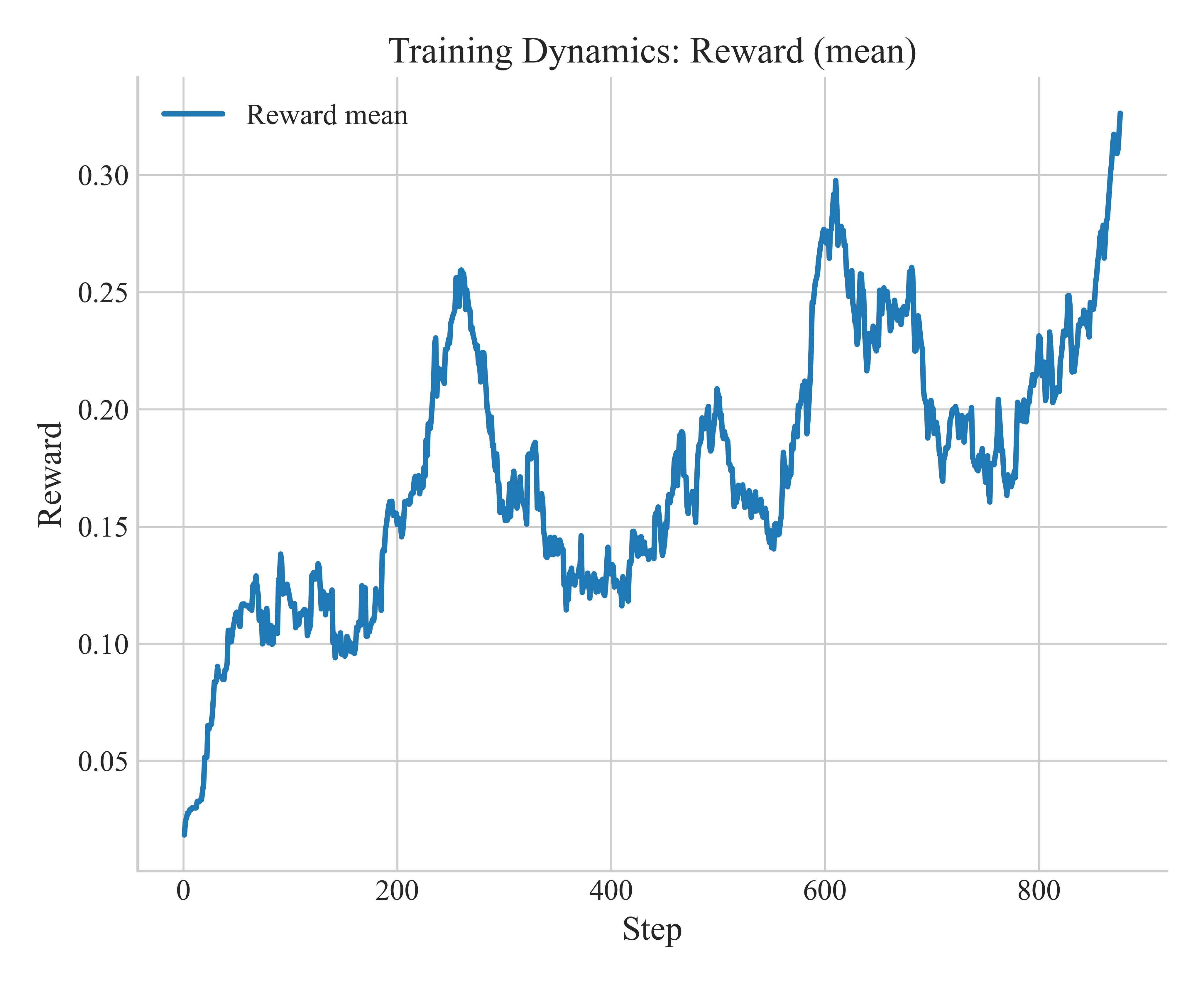}
        \caption{Training dynamics of the mean reward under the proposed information gain.}
    \end{subfigure}
    \caption{Training dynamics of DAPO under the information gain reward objective.}
    \label{fig:training_dynamics}
\end{figure*}

\paragraph{On-Policy Monte Carlo Estimation.} Our training process can be interpreted as a Monte Carlo estimation of the expected Bayesian utility.
The policy $\pi_\theta(\cdot|x_t)$ functions as a proposal distribution, sampling a group of $K$ candidate questions $\{Q^{(k)}\}_{k=1}^K$, and $A^{(k)}$ is the response returned by the strict user simulator for the sampled question $Q^{(k)}$. The empirical average of their rewards provides an unbiased estimate of the expected utility:
\begin{equation}
    \mathcal{U}(\theta) \approx \mathbb{E}_{Q \sim \pi_\theta} [R_t] \approx \frac{1}{K} \sum_{k=1}^{K} R_t(x_t, Q^{(k)},A^{(k)}).
\end{equation}
As training progresses, $\pi_\theta$ increasingly concentrates probability mass on the regions of the question space that yield high information gain.

\paragraph{Advantage Computation.}
To reduce variance and stabilize training, we compute the group-relative advantage for each candidate $k$:
\begin{equation}
    Adv_k = \frac{R_t^{(k)} - \mu_K({R_t})}{\sigma_K({R_t}) + \epsilon},
\end{equation}
where $\mu_K$ and $\sigma_K$ denote the mean and standard deviation across the sampled group, and $\epsilon$ is a small constant for numerical stability. Since $R_t$ is a sequence-level metric derived from the final belief update, we broadcast the scalar advantage $Adv_k$ to all tokens within the question $Q^{(k)}$. 
 \paragraph{DAPO Objective.}
We optimize the policy using the decoupled clipping objective, which is particularly effective for the sparse reward nature of clarification tasks:
\begin{equation}
\small
\begin{split}
\mathcal{J}(\theta) &= \frac{1}{\sum_{k} |Q^{(k)}|} \sum_{k=1}^{K} \sum_{j=1}^{|Q^{(k)}|} \min \Big( \rho_{k,j} Adv_k, \\
&\quad \text{clip}(\rho_{k,j}, 1-\epsilon_{\text{low}}, 1+\epsilon_{\text{high}}) Adv_k \Big)
\end{split}
\label{eq:dapo_loss}
\end{equation}
where $\rho_{k,j}$ is the importance sampling ratio. The clipping function restricts $\rho_{k,j}$ within $[1-\epsilon_{\text{low}}, 1+\epsilon_{\text{high}}]$ (with $\epsilon_{\text{high}} > \epsilon_{\text{low}}$), effectively raising the ceiling for exploration on high-potential tokens. By maximizing this objective, the clarifier learns to align its clarifying behavior with the goal of maximizing information gain, effectively amortizing the cost of optimal experimental design.

\section{Experiments}
\label{sec:exp_setup2}

\begin{table*}[t]
\centering
\small
\caption{
Ablation study on $\tau$-Bench (\texttt{Retail} and \texttt{Airline}) based on training with Qwen3-1.7B.
Pass@1 (\%) with deterministic decoding (temperature $=0$), averaged over three runs. Numbers in parentheses indicate changes relative to \textit{None}. \textit{Baseline} is the untrained Qwen3-1.7B clarifier; \textit{LLM as a Judge} replaces the information-gain reward with a question-quality judge; \textit{w/o Information Gain} optimizes the posterior likelihood term without the prior-posterior contrast.
}
\setlength{\tabcolsep}{8.6pt}
\begin{tabular}{lcccccc}
\toprule
\multicolumn{1}{c}{\textbf{Method}} 
& \multicolumn{3}{c}{\textbf{Success Rate (\% $\uparrow$)}} 
& \multicolumn{3}{c}{\textbf{Steps $\downarrow$}} \\
\cmidrule(lr){1-1}
\cmidrule(lr){2-4}
\cmidrule(lr){5-7}
\textbf{Strategy} 
& \textbf{Retail} 
& \textbf{Airline} 
& \textbf{Average} 
& \textbf{Average (w/o Clarifier)} 
& \textbf{Average Clarify} 
& \textbf{Sum} \\
\midrule
None 
& 16.5 
& 14.3 
& 15.4 
& 24.2
& 0.0 
& 24.2 \\
\midrule
Baseline (Qwen3-1.7B)
& 16.5 \textcolor{gray}{\scriptsize(+0.0)} 
& 13.3 \textcolor{gray}{\scriptsize(-1.0)} 
& 14.9 \textcolor{gray}{\scriptsize(-0.5)} 
& 23.9
& 4.2 
& 28.1 \\
LLM as a Judge
& \textbf{18.3} \textcolor{gray} {\scriptsize(+1.8)} 
& 8.0 \textcolor{gray} {\scriptsize(-6.3)} 
& 13.2 \textcolor{gray} {\scriptsize(-2.2)} 
& 26.7 
& 2.9 
& 29.6 \\
w/o Information Gain 
& 15.6 \textcolor{gray} {\scriptsize(-0.9)} 
& 10.0 \textcolor{gray} {\scriptsize(-4.3)} 
& 12.8 \textcolor{gray} {\scriptsize(-2.6)} 
& 26.8 
& 2.4 
& 29.2 \\
\rowcolor{blue!10}
Ours 
& \textbf{18.3} \textcolor{gray} {\scriptsize(+1.8)} 
& \textbf{17.3} \textcolor{gray} {\scriptsize(+3.0)} 
& \textbf{17.8} \textcolor{gray} {\scriptsize(+2.4)} 
& 23.8 
& 1.3 
& 25.1 \\
\bottomrule
\end{tabular}
\vspace{3pt}
\footnotesize

\textit{Notes.}
$\uparrow$ ($\downarrow$) indicates higher (lower) is better.
\textit{Average} denotes the mean across domains.
\textit{Average (w/o Clarifier)} counts agent and user steps excluding clarifier turns;
\textit{Average Clarify} denotes the average number of clarifier invocations;
\textit{Sum} counts all \textit{Average (w/o Clarifier)} and \textit{Average Clarify} steps.
w/o Information Gain removes the uncertainty reduction term and optimizes
$\log P(G \mid x_t, Q_t, A_t)$ only.
\label{tab:clarifier-passk}
\end{table*}

Building upon the framework introduced in Section~\ref{method}, we evaluate our approach within the clarifier-augmented $\tau$-Bench environment. To ensure a clean separation between training and evaluation, all models are fine-tuned \emph{exclusively} on a dataset of 500 $\tau$-retail training trajectories. Testing is subsequently conducted on two distinct sets: 115 held-out \texttt{retail} tasks, which represent the in-domain setting, and 50 \texttt{airline} tasks, which represent the out-of-distribution (OOD) setting. The \texttt{airline} domain serves as a particularly challenging testbed for transfer learning, as it features task structures, tool APIs, and user interaction patterns that differ markedly from those observed during training. Further implementation details and specific training hyperparameters are provided in Appendix~\ref{hy}.
\begin{table*}[t]
\centering
\small
\caption{ 
Exploration Boundaries of Agents and Clarifier: Impact of Different Clarification Strategies on Task Completion Performance.  
Numbers in parentheses indicate changes relative to the agent's performance under the \textit{None} condition (fuzzy user intent without clarification). $\uparrow$ ($\downarrow$) indicates higher (lower) is better.  
\textit{None (Full User Intent)} represents the condition where full user intent is provided without clarification, reflecting the agent's performance boundaries.
}
\setlength{\tabcolsep}{8pt}
\begin{tabular}{lcccccc}
\toprule
\multicolumn{1}{c}{\textbf{Method}} 
& \multicolumn{3}{c}{\textbf{Success Rate (\% $\uparrow$)}} 
& \multicolumn{3}{c}{\textbf{Steps $\downarrow$}} \\
\cmidrule(lr){1-1} \cmidrule(lr){2-4} \cmidrule(lr){5-7}
\textbf{Clarify Strategy} 
& \textbf{Retail} 
& \textbf{Airline} 
& \textbf{Average} 
& \textbf{Average (w/o Clarifier)} 
& \textbf{Average Clarify} 
& \textbf{Sum} \\
\midrule
None 
& 16.5 
& 14.3 
& 15.4 
& 24.2
& 0.0 
& 24.2 \\
\midrule
Qwen3-8B  
& 15.7 \scriptsize{(-0.8)} 
& 8.0 \scriptsize{(-6.3)} 
& 11.9 \scriptsize{(-3.5)} 
& 23.1
& 2.7 
& 25.8
\\
Qwen3-14B  
& 16.5 \scriptsize{(+0.0)} 
& 16.0 \scriptsize{(+1.7)} 
& 16.3 \scriptsize{(+0.9)} 
& 24.2
& 5.1
& 29.3
\\
Qwen3-30B  
& 15.7 \scriptsize{(-0.8)} 
& \textbf{19.0} \scriptsize{(+4.7)} 
& 17.4 \scriptsize{(+2.0)} 
& 21
& 3.8
& 24.8
\\
DeepSeek-R1  
& 14.8 \scriptsize{(-1.7)} 
& \textbf{19.0} \scriptsize{(+4.7)} 
& 16.9 \scriptsize{(+1.5)} 
& 20.5
& 3.9
& 24.4
\\
DeepSeek-V3.1  
& \textbf{19.1} \scriptsize{(+2.6)} 
& 17.0 \scriptsize{(+2.7)} 
& \textbf{18.1} \scriptsize{(+2.7)} 
& 21
& 3.9
& 24.9
\\
\rowcolor{blue!10}
Ours (Qwen3-1.7B-trained)  
& 18.3 \textcolor{gray} {\scriptsize(+1.8)} 
& 17.3 \textcolor{gray} {\scriptsize(+3.0)} 
& 17.8 \textcolor{gray} {\scriptsize(+2.4)} 
& 23.8 
& 1.3 
& 25.1 \\
\midrule
None (Full User Intent)
& 20 
& 16
&  18
& 22.3
& 0.0 
& 22.3 \\
\bottomrule
\end{tabular}
\vspace{3pt}
\footnotesize
\label{tab:clarifier-comparison}
\end{table*}

\subsection{Training Dynamics and Performance}
We first analyze the training dynamics to assess whether optimization steers the agent toward uncertainty-reducing behaviors. Figure~\ref{fig:training_dynamics}(a) decomposes our Information Gain Reward into two teacher-forced log-likelihood terms: the prior confidence in the ground-truth goal given the interaction context alone, $\log P(G^* \mid x)$, and the posterior confidence after incorporating the clarification exchange, $\log P(G^* \mid x, Q, A)$. Notably, the likelihood conditioned on the interaction context without clarification, $\log P(G^* \mid x)$, remains mildly oscillatory throughout training, reflecting variation in sampled contexts during on-policy training. However, $\log P(G^* \mid x)$ remains substantially below the $\log P(G^* \mid x, Q, A)$ posterior curve, and the gap between the two curves tends to widen with training. This pattern suggests that the learned policy improves primarily by eliciting additional task-relevant evidence, not by simply increasing prior confidence in ambiguous contexts.

Consistent with this interpretation, Figure~\ref{fig:training_dynamics}(b) shows that the mean Information Gain Reward increases steadily over training, rising from a near-zero baseline to approximately 0.15 by step 200 and continuing to improve thereafter. This monotonic trend, without signs of collapse or oscillatory behavior, suggests that the reward provides a stable optimization signal. Together with the divergence observed in Figure~\ref{fig:training_dynamics}(a), these results indicate that DAPO effectively exploits the information gain objective to refine the clarification policy, progressively improving the agent’s ability to acquire informative feedback.
\begin{figure*}
    \centering
    \begin{subfigure}[t]{0.493\linewidth}
        \centering
        \includegraphics[width=\linewidth]{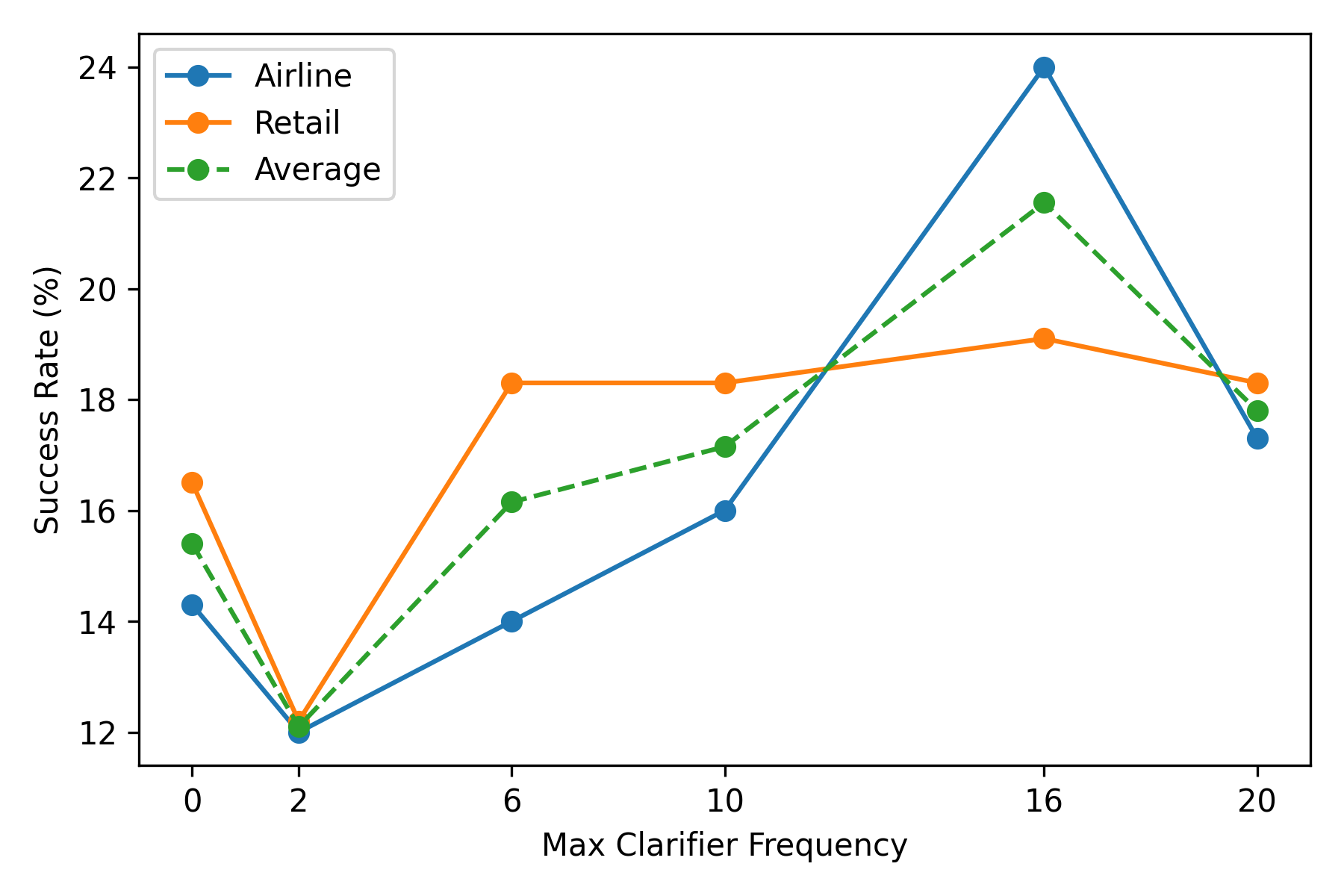}
        \caption{Success rate on the \texttt{airline} and \texttt{retail} domains, and their average, under different maximum clarification budgets.}
        \label{fig:pass_vs_clarify}
    \end{subfigure}
    \hspace{0.000\linewidth}
    \begin{subfigure}[t]{0.493\linewidth}
        \centering
        \includegraphics[width=\linewidth]{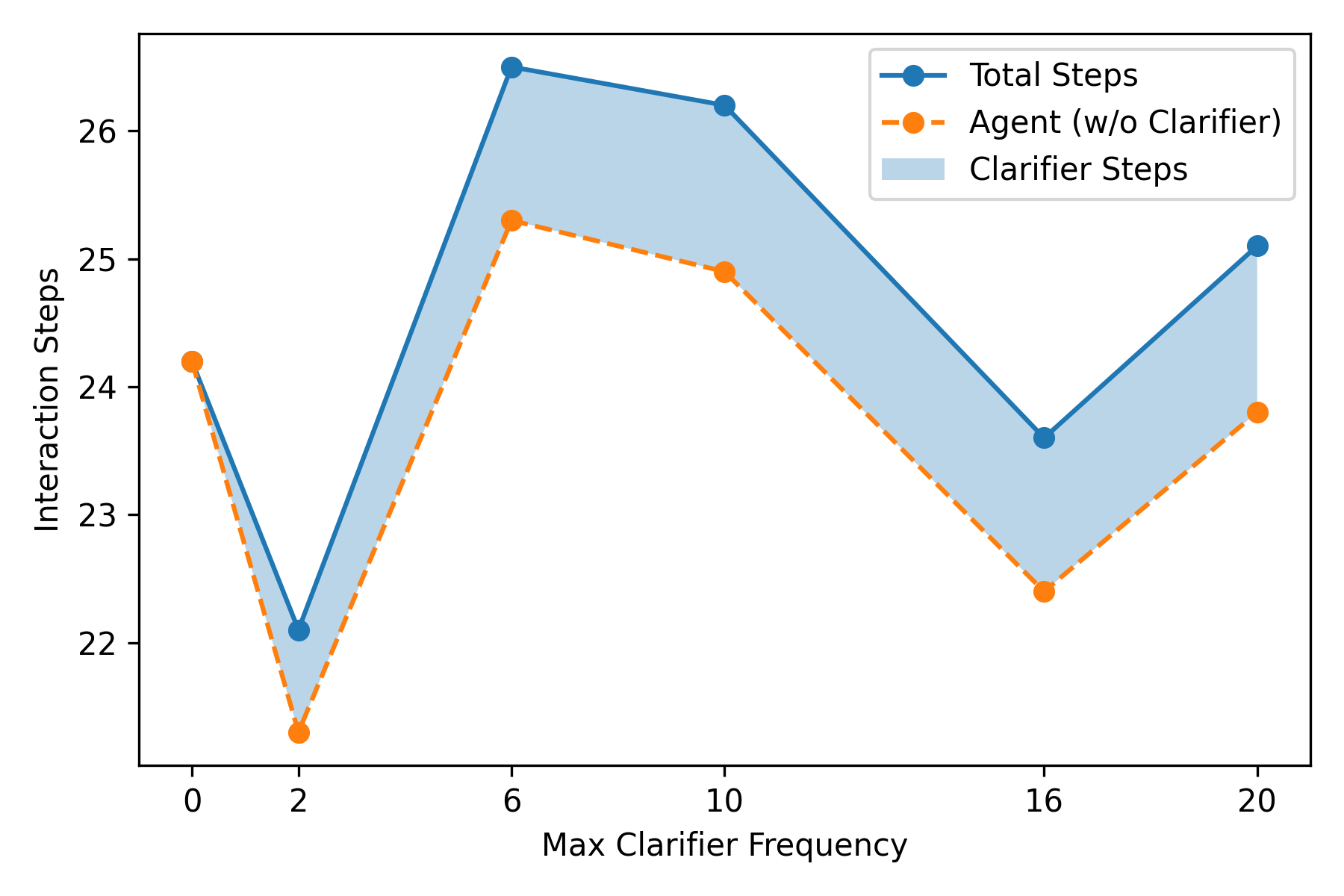}
         \caption{Average number of interaction steps, including clarification steps (shaded area), across the \texttt{airline} and \texttt{retail} domains.}
        \label{fig:steps_vs_clarify}
    \end{subfigure}
    \caption{
        Effect of clarification budget on task success and interaction efficiency under forced sampling.
    }

    \label{fig:clarification_tradeoff}
\end{figure*}

Table~\ref{tab:clarifier-passk} reports an ablation study comparing different clarification training strategies on $\tau$-Bench.  Compared to the pre-trained Qwen3-1.7B clarifier, our method consistently improves success rates on both domains, increasing performance from 16.5\% to 18.3\% on \texttt{retail} and from 13.3\% to 17.3\% on \texttt{airline}, while simultaneously reducing interaction cost. In particular, the average number of clarifier invocations drops sharply from 4.2 to 1.3 per task, resulting in a 3 reduction in total interaction steps, indicating that the learned policy asks fewer but more targeted clarification questions.  When the available information is already sufficient, the prior belief over user intent is already highly concentrated. In this case, additional clarification often introduces weakly informative or irrelevant context, making the prediction of the ground-truth action sequence under teacher forcing more difficult and leading to a decrease in log-likelihood. Intuitively, unnecessary questions introduce distracting information without providing meaningful constraints on the underlying intent, and are therefore less favorable than abstention (reward = 0).

In contrast, removing the information gain term and optimizing with a likelihood-only objective (w/o Information Gain) leads to a substantial performance degradation, especially on the \texttt{airline} domain, where success rate falls from 17.3\% to 10\%, suggesting that conditional likelihood alone is insufficient to encourage informative clarifications in domains.
The LLM-as-a-judge variant further exhibits unstable behavior: although it matches our method on \texttt{retail} (18.3\%), performance collapses on \texttt{airline} (8\%) and incurs higher interaction costs, highlighting the limitation of stylistic or preference-based supervision in capturing the task-specific informational value of clarification questions. Details of the LLM-as-a-judge prompt template are provided in Appendix~\ref{LLM-as-judge}.

\subsection{Exploration of Clarifier Boundaries}
Table~\ref{tab:clarifier-comparison} compares our specialized 1.7B clarifier against a wide range of large-scale LLMs (from 8B to 671B). Despite the significant disparity in parameter scale, our method achieves a task success rate of 17.8\%, approaching the best-performing large backbone (DeepSeek-V3.1, 18.1\%) with a negligible margin of merely \textbf{0.3\%}. This result demonstrates that a small, information-driven model can effectively match the utility of significantly larger models in resolving task-oriented ambiguity.

Crucially, this performance is achieved with superior interaction efficiency. While general-purpose models tend to over-clarify with 3.9 to 5.1 turns per task, our policy converges with just \textbf{1.3} turns on average. This efficient strategy is vital for real-world deployment, as it mitigates the deleterious effects of over-clarification, including interactional redundancy, user fatigue, and the risk of task state pollution (where excessive dialogue history distracts the agent). By optimizing for Information Gain, our clarifier strikes a favorable balance, facilitating ambiguity resolution while reducing the cognitive and temporal costs of interaction.

Comparing against the \textit{Full User Intent} offers a profound insight into information modality (see Appendix~\ref{Full-user-prompts} for the specific prompt template). While this static oracle serves as an upper bound of agent's capability, our interactive approach surprisingly outperforms it in the complex \texttt{Airline} domain (17.3\% vs. 16\%). This suggests that in intricate scenarios, interactive ambiguity resolution is superior to static information loading. By breaking down complex constraints into sequential turns, our clarifier mitigates the risk of information overload, enabling the agent to attend to task-critical details more effectively than when overwhelmed by a monolithic instruction dump. Overall, our learned clarifier effectively approximates this theoretical performance boundary (17.8\% vs. 18\% on average) while incurring a marginal overhead of only 2.8 total steps.

\begin{table*}[t]
\centering
\small
\setlength{\tabcolsep}{7.6pt}
\caption{
Agent and clarifier generalization performance with the clarifier decoded deterministically. Different agent backbones are compared under varying clarification strategies, including no clarification (\textit{None}), a \textit{Qwen3-8B} clarifier, and a learned 1.7B clarifier (\textit{Ours}) trained with our DAPO setting. $\uparrow$ ($\downarrow$) indicates higher (lower) is better. The best results for each model are bolded. The values in parentheses indicate the change relative to the performance with no clarification None.
}
\begin{tabular}{llcccccc}
\toprule
\multicolumn{2}{c}{\textbf{Setting}} & \multicolumn{3}{c}{\textbf{Success Rate (\%$\uparrow$)}} & \multicolumn{3}{c}{\textbf{Steps $\downarrow$}} \\
\cmidrule(r){1-2} \cmidrule(lr){3-5} \cmidrule(l){6-8}
\textbf{Agent Model} 
& \textbf{Clarifier Model}
& \textbf{Retail} 
& \textbf{Airline} 
& \textbf{Average} 
& \textbf{Average (w/o Clarifier)} 
& \textbf{Clarify} 
& \textbf{Sum} \\
\midrule
\multirow{3}{*}{Qwen3-8B}  & None  & 16.5& 14.3 & 15.4 & 24.2& 0 & 24.2  \\
 & Qwen3-8B & 15.7 {\scriptsize(-0.8)} & 8.0 {\scriptsize(-6.3)} & 11.9 {\scriptsize(-3.5)} & 23.1 & 2.7 & 25.8\\
 \rowcolor{blue!10} & Ours & \textbf{18.3} {\scriptsize(+1.8)} &\textbf{17.3} {\scriptsize(+3.0)}& \textbf{17.8} {\scriptsize(+2.4)} &23.8 & 1.3 & 25.1\\
\midrule
\multirow{3}{*}{Qwen3-32B}  & None & 23.4 & 14 & 18.7 & 24 & 0& 24 \\
& Qwen3-8B &21.7 {\scriptsize(-1.7)} & 10 {\scriptsize(-4.0)}& 15.9 {\scriptsize(-2.6)}& 20.4 &5.1 & 25.5 \\
 \rowcolor{blue!10}& Ours & \textbf{24.3} {\scriptsize(+0.9)}  & \textbf{18} {\scriptsize(+4.0)} & \textbf{21.2} {\scriptsize(+2.5)} & 16.9 & 2.6 & 19.5\\
\midrule

\multirow{3}{*}{DeepSeek-R1}  & None & 31.3 & 40 & 35.7 & 18.1 & 0 & 18.1 \\
 & Qwen3-8B & 26.0 {\scriptsize(-5.3)}   & 42 {\scriptsize(+2.0)}  &  34 {\scriptsize(-1.7)} & 11.5 & 1 & 12.5 \\
 \rowcolor{blue!10} & Ours & \textbf{33.9} {\scriptsize(+2.6)} & \textbf{46} {\scriptsize(+6.0)}& \textbf{40} {\scriptsize(+4.3)}&21.8 & 0.5  & 22.3\\
\midrule{}
\multirow{3}{*}{DeepSeek-V3.2}& None  &36.5 & 30 & 33.3&23.5&0 & 23.5  \\
& Qwen3-8B  &\textbf{39.1} {\scriptsize(+2.6)} & 40 {\scriptsize(+10)} & 39.6 {\scriptsize(+6.3)} & 23.3 & 2.7& 26\\
 \rowcolor{blue!10}& Ours  &37.4 {\scriptsize(+0.9)} &\textbf{42} {\scriptsize(+12)}&\textbf{39.7} {\scriptsize(+6.4)} & 23.1& 0.7 &23.8 \\
\midrule
\multirow{3}{*}{GLM-4-32B}  & None &6.1 &24& 15 & 33.3 & 0 &33.3 \\
& Qwen3-8B & \textbf{7} {\scriptsize(+0.9)} & 20 {\scriptsize(-4.0)} & 13.5 {\scriptsize(-1.5)} & 35.6 & 1.5 & 37.1 \\
 \rowcolor{blue!10}& Ours &\textbf{7} {\scriptsize(+0.9)} & \textbf{26} {\scriptsize(+2.0)} & \textbf{16.5} {\scriptsize(+1.5)} & 32.9 & 1 & 33.9 \\
\midrule
\multirow{3}{*}{Agent Avg.} & None & 22.8 & 24.5 & 23.6 &24.6 & 0 & 24.6\\
 & Qwen3-8B & 21.9 {\scriptsize(-0.9)} & 24 {\scriptsize(-0.5)} & 23 {\scriptsize(-0.6)}& 22.8 & 2.6 & 25.6 \\
 \rowcolor{blue!10} & Ours & \textbf{24.2} {\scriptsize(+1.4)} & \textbf{29.9} {\scriptsize(+5.4)} & \textbf{27.3} {\scriptsize(+3.7)} &23.7 & 1.2 & 24.9 \\

\bottomrule
\end{tabular}
\label{tab:agent-gen}
\end{table*}


\subsection{Impact of Clarification Trigger Frequency}

Figure~\ref{fig:clarification_tradeoff}(a) examines how the maximum clarification budget affects Pass@1 success rates on the \texttt{airline} and \texttt{retail} domains. Introducing clarification consistently improves performance over the no-clarification setting, demonstrating its effectiveness in resolving underspecified user intents. Increasing the budget from 0 to a moderate level yields a substantial gain in average success rate, rising from approximately 15\% to over 21\%. However, the relationship between clarification budget and performance is clearly non-monotonic: success rates peak at intermediate budgets and plateau or decline as the budget increases further, with the effect being most pronounced in the \texttt{airline} domain. This finding challenges the common design assumption that more clarification is inherently beneficial. Instead, it suggests that excessive clarification introduces context pollution and redundant interactions, which can disrupt the agent's reasoning flow without adding task-relevant constraints.

Figure~\ref{fig:clarification_tradeoff}(b) reveals a relationship between clarification frequency and task efficiency. Initially, increasing the budget from 0 to 2 reduces total interaction steps from 24.2 to 22.1 without improving the success rate, indicating that insufficient questioning fails to resolve core task ambiguities. A significant efficiency gain occurs as the budget scales from 6 to 16: while the success rate grows consistently, total steps peak at 26.5 (budget 6) before declining to 23.6 (budget 16). This trend demonstrates that effective mid-stage clarification successfully converts into substantial performance gains while streamlining the execution path by preempting redundant tool actions. Beyond budget 16, interaction steps rise to 25.1 with no commensurate success rate gains. This reversal suggests that excessive clarification becomes counterproductive, introducing informational noise that distracts the agent rather than facilitating goal resolution. To further understand the remaining failure modes, we conduct a qualitative audit of representative trajectories and distinguish clarification failures from downstream execution errors; detailed analyses and case studies are provided in Appendix~\ref{appendix:failure_cases}.

\subsection{Cross-Agent Generalization Ability}
Table~\ref{tab:agent-gen} presents the cross-agent generalization performance under deterministic decoding across a diverse set of agents, including Qwen3-32B, DeepSeek-R1, DeepSeek-V3.2, and GLM-4-32B. We evaluate three clarification strategies: no clarification (\textit{None}), a publicly released model (\textit{Qwen3-8B}) as a clarifier, and our learned clarifier (\textit{Ours}), which is trained using information-gain-driven DAPO.
\paragraph{Overall trends.} Averaged across all five agent backbones, our method consistently improves task success relative to the baseline. \textit{Ours} achieves the highest average success rate (27.3\%), outperforming both \textit{None} (23.6\%) and the Qwen3-8B clarifier (23\%). In particular, \textit{Qwen3-8B} fails to generalize effectively, resulting in a slight regression in average performance (-0.6\%), whereas \textit{Ours} delivers a robust gain of 3.7\%. This divergence is most pronounced in the out-of-distribution \texttt{Airline} domain, where our method boosts success by 5.4\%, confirming that explicitly resolving underspecified user intent improves performance across heterogeneous downstream agents.

\paragraph{Comparison across clarification strategies.} The \textit{Qwen3-8B} clarifier exhibits high variance: while it benefits specific models like DeepSeek-V3.2, it adversely impacts others, such as Qwen3-32B and GLM-4-32B. In contrast, \textit{Ours} delivers stable positive outcomes across all backbones while maintaining a significantly lower clarification frequency (1.2 vs. 2.6 turns on average). This efficiency stems from a policy trained to selectively intervene only when clarification is expected to significantly reduce information asymmetry, effectively avoiding the redundant or disruptive questioning.

\paragraph{Agent-level success rate analysis.} We observe a clear positive correlation between the agent's intrinsic capability and the magnitude of improvement. For mid-sized agents like Qwen3-32B, our approach improves the average success rate from 18.7\% to 21.2\% (+2.5\%). Notably, these benefits scale significantly with stronger models: the larger scale DeepSeek-R1 sees a robust boost from 35.7\% to 40\% (+4.3\%), while the massive 671B-parameter DeepSeek-V3.2 achieves the most substantial improvement, rising from 33.3\% to 39.7\% (+6.4\%). This trend suggests that more capable agents are better equipped to leverage the high-quality additional constraints provided by our clarifier, effectively unlocking their full potential in ambiguous scenarios without disrupting their intrinsic reasoning processes. We further explore that the proposed training framework generalizes across different clarifier backbones, including Qwen2.5-1.5B and Qwen3-4B; detailed results are provided in Appendix~\ref{appendix:backbone_generalization}.

\paragraph{Interaction efficiency across agents.} We find that improved task performance does not necessarily come with increased interaction costs. For agents like Qwen3-32B and DeepSeek-V3.2, clarification leads to a net reduction in total steps (e.g., 24 to 19.5 for Qwen3-32B), suggesting that early ambiguity resolution effectively preempts longer, erroneous execution trajectories. On average, across all agent backbones, our method adds just 1.2 clarification turns and 0.3 total steps per task, demonstrating a surgical intervention strategy that achieves favorable success--cost trade-offs.

\section{Conclusion}

We present an information-gain-driven Clarifier-augmented framework to optimize the timing and content of questioning. Through a comprehensive evaluation spanning training dynamics, clarifier boundaries, trigger frequency, and cross-agent generalization, we demonstrate the robustness of our method across diverse settings. Empirical results show that our approach improves the average success rate by 3.7\% over baselines with minimal overhead (averaging 0.3 additional steps), demonstrating the effectiveness of information gain for efficient ambiguity resolution. 

\textbf{Discussion and Future Work.} In practical agent systems, clarification should be treated as a selectively routed capability rather than an always-on dialogue behavior. A base agent or external controller needs to determine whether the current state contains unresolved intent uncertainty, whether clarification is permitted under domain-specific policies, and whether the expected informational benefit outweighs the associated user burden. Our current setup also leaves several directions for future work. The training reward relies on access to ground-truth user goals in controlled environments, while real users may provide noisy, incomplete, or ambiguous feedback over multiple turns. Future work could therefore explore weaker supervision signals and more realistic interaction settings. In addition, the current experiments keep the base agent fixed for clearer attribution of clarification behavior, whereas jointly optimizing the agent and clarifier may further improve coordination between action selection and information seeking.

\section*{Acknowledgement}

This work is supported by Advanced Materials-National Science and Technology Major Project (Grant No. 2025ZD0620100), HKUST(GZ)-IEIP-RoP (G01RF000256),  National Key R\&D Program of China (No. 2024YFA1012700), and Guangdong Provincial Key Lab of Integrated Communication, Sensing and Computation for Ubiquitous Internet of Things (No. 2023B1212010007).
\section*{Impact Statement}
This paper presents work aimed at improving the reliability and efficiency of tool-using LLM agents through a Bayesian Information Gain framework. Our primary goal is to reduce ambiguity in human-agent interaction, thereby preventing erroneous tool executions. Our method focuses on optimizing clarification strategies within controlled environments and we do not foresee any specific negative societal consequences or ethical concerns that require immediate highlighting, as the proposed method operates strictly within the scope of user-initiated tasks and grounded tool definitions.

\bibliography{example_paper}
\bibliographystyle{icml2026}
\clearpage
\onecolumn
\section{Appendix}

\newtcolorbox{promptbox}{
  colback=gray!5,
  colframe=gray!50,
  boxrule=0.5pt,
  arc=2pt,
  left=5pt,
  right=5pt,
  top=5pt,
  bottom=5pt,
  fontupper=\ttfamily\footnotesize,
  enhanced,
  breakable 
}

\subsection{Data Construction}
\label{app:data-construct}
Our data construction pipeline focuses on collecting interaction trajectories from a Clarifier-augmented $\tau$-Bench environment and extracting step-level interaction histories for training the clarification policy via DAPO. Unlike prior approaches that rely on supervised labels or preference annotations, our method does not use SFT or DPO data. Instead, learning is driven entirely by on-policy rollouts and reward signals defined over interaction histories.

\paragraph{Interaction Trajectory Generation.}
All interaction trajectories are collected from the Clarifier-augmented environment described in Section~\ref{sec:env}. During data collection, the downstream task agent is Qwen3-8B, which interacts with domain-specific tools under the standard $\tau$-Bench execution protocol. A separate Clarifier module, implemented using Qwen3-32B, is invoked only after tool calls to determine whether clarification is necessary and, if so, to generate a follow-up question grounded in the observed execution feedback. User responses to clarification questions are produced by an LM-simulated user instantiated as Qwen3-8B.

This setup yields multi-turn interaction trajectories of the form
\[
\tau = \{\,U,\; (a_t, o_t, Q_t, A_t)_{t=1}^{T},\; G\,\},
\]
where $U$ denotes the initial user instruction, $(a_t, o_t)$ are the agent’s tool actions and corresponding observations, $(Q_t, A_t)$ are optional clarification question--answer pairs, and $G$ is the task-defined ground-truth user goal provided by $\tau$-Bench. Clarification turns are recorded only when triggered; otherwise, $Q_t = A_t = \varnothing$.

\paragraph{Step-Level Instance Extraction.}
From the collected trajectories, we construct training instances at the step level. Each instance corresponds to a decision point $t$ and consists solely of the interaction history
\[
x_t = \big(U,\; (a_i, o_i, Q_i, A_i)_{i < t}\big),
\]
together with the associated ground-truth goal $G$. We restrict our scope to the first 16 agent steps of each trajectory to emphasize ambiguity resolution, where clarification is most critical. Importantly, we do not retain or supervise on the clarifier outputs generated during data collection; all clarification decisions and questions used for training are produced by the Qwen3-1.7B model during on-policy rollouts.

\paragraph{From $G$ to $G^*$.}
A critical component of our pipeline for reward learning is the normalization of the user goal. The raw goals in $\tau$-Bench often contain extraneous personality descriptions (e.g., ``logical,'' ``shy'') and complex JSON syntax that can distract smaller models. To address this, we employ an LLM-based rewriting step to transform the raw user goals into a flat, natural language format for efficient supervision. The rewriting strictly enforces the following rules:
\begin{itemize}
    \item \textbf{First-person normalization}: converting third-person identifiers to first-person intent (e.g., ``Your name is'' to ``My name is'').
    \item \textbf{Structure flattening}: transforming JSON key--value pairs into cohesive natural language phrases.
    \item \textbf{Noise removal}: removing non-functional attributes while strictly preserving specific values such as addresses and IDs.
\end{itemize}

This process yields a clean, unambiguous normalized goal $G^*$, which serves as the supervision target for reward computation throughout this paper. The detailed prompt template and representative rewriting examples are provided in Appendix~\ref{app:data_prompt}. 

\subsection{Derivation of the Information Gain Reward}
\label{app:proof}

In Section~\ref{sec:reward}, we defined the reward $R_t$ as the difference between the posterior and prior log-likelihoods of the ground-truth goal $G^*$. Here, we derive its connection to the Bayesian Importance Weight.

Let $x$ be the context, $Q$ be the clarification question, and $A$ be the user's response. The reward is given by:
\begin{equation}
    R_t = \log P(G^* \mid x, Q, A) - \log P(G^* \mid x).
\end{equation}
Using Bayes' theorem, we can expand the posterior term $P(G^* \mid x, Q, A)$:
\begin{equation}
    P(G^* \mid x, Q, A) = \frac{P(A \mid G^*, Q, x) P(G^* \mid x, Q)}{P(A \mid Q, x)}.
\end{equation}
Since the goal $G^*$ exists independently of the question $Q$ (the user's intent is fixed before the question is asked), we have $P(G^* \mid x, Q) = P(G^* \mid x)$. Substituting this back into the equation:
\begin{equation}
    P(G^* \mid x, Q, A) = \frac{P(A \mid G^*, Q, x) P(G^* \mid x)}{P(A \mid Q, x)}.
\end{equation}
Now, substituting this expression into the reward definition:
\begin{equation}
    \begin{aligned}
    R_t &= \log \left( \frac{P(A \mid G^*, Q, x) P(G^* \mid x)}{P(A \mid Q, x)} \right) - \log P(G^* \mid x) \\
                &= \log \left( \frac{P(A \mid G^*, Q, x)}{P(A \mid Q, x)} + \log P(G^* \mid x) - \log P(G^* \mid x)  \right) \\
        &= \log \left( \frac{P(A \mid G^*, Q, x)}{P(A \mid Q, x)} \right).
    \end{aligned}
\end{equation}
The term inside the log is the Importance Weight $w$:
\begin{equation}
    w = \frac{P(A \mid G^*, Q, x)}{P(A \mid Q, x)}.
\end{equation}
This ratio compares the likelihood of the answer $A$ under the specific ground-truth goal $G^*$ versus its likelihood under the general marginal distribution. A high weight $w$ (and thus high $R_t$) implies that the answer $A$ is highly specific to $G^*$ and unlikely to occur by chance, thereby providing strong discriminative evidence.

\subsection{Case Studies and Prompt Templates}
\subsubsection{Reward Case Studies}
\label{reward_case}

We present a detailed comparison of two clarification trajectories from the same interaction state. This case study illustrates how the Information Gain Proxy ($R_t$) effectively penalizes redundant dialogue while rewarding interactions that actively elicit missing constraints required by the ground truth.

\noindent\textbf{Ground Truth User Goal ($G^*$):}
\begin{quote}
\small
My name is Olivia Lopez and my zip code is 76171. For \#W7449508, exchange Sneakers size 6, black, synthetic to size 10, gray, leather; via gift\_card\_7711863.
\end{quote}

Table~\ref{tab:detailed_reward_comparison} contrasts a redundant confirmation loop against a high-quality clarification turn.

\begin{table*}[h]
    \centering
    \small
    \renewcommand{\arraystretch}{1.5}
    \caption{Detailed comparison of clarification candidates. \textbf{Case B} demonstrates a high-quality question that specifically targets the missing payment information. This targeted inquiry successfully recovers the \texttt{gift\_card} detail, leading to a significant increase in the ground truth's log-probability.}
    \label{tab:detailed_reward_comparison}
    
    \begin{tabular}{p{0.15\textwidth} | p{0.40\textwidth} | p{0.40\textwidth}}
        \toprule
        \textbf{Component} & \textbf{Case A: Low Information Gain} & \textbf{Case B: High Information Gain} \\
        \midrule
        \textbf{Question ($Q_t$)} & Can you confirm if you still want to proceed with exchanging your item for the gray, leather sneakers in size 10, as the system indicates it's available? & To finalize the exchange for the gray leather sneakers in size 10, could you please provide the payment method you wish to use? \\
        \midrule
        \textbf{User Answer ($A_t$)} & Yes, I want to proceed with the exchange for the gray, leather sneakers in size 10. That's what I asked for. & I need to exchange my black, synthetic sneakers size 6 for gray, leather sneakers size 10. Can you do that with gift\_card\_7711863? \\
        \midrule
        \textbf{Analysis} & \textit{Redundant Confirmation}: The user re-states known preferences. No new entities are introduced. & \textit{Targeted Elicitation}: The agent identifies the missing payment slot and asks for it directly. The user provides the gift card ID. \\
        \midrule
        \textbf{Log Prob ($x_t$)} & \multicolumn{2}{c}{$-3.2649$} \\
        \midrule
        \textbf{Log Prob ($x_t,Q,A$)} & $-3.3193$ \quad (\%$\downarrow$) & $\mathbf{-2.6153}$ \quad (\%$\uparrow$) \\
        \midrule
        \textbf{Reward ($R_t$)} & $-0.0543$ & $\mathbf{0.6496}$ \\
        \bottomrule
    \end{tabular}
\end{table*}

\noindent\textbf{Mechanism Discussion:}

\begin{itemize}
    \item \textbf{High Information Gain (Case B):} The agent asks a high-quality question targeting the missing \textit{payment method}. This prompts the user to reveal the token sequence ``\texttt{gift\_card\_7711863}''. Since this specific ID is a mandatory part of the Ground Truth $G^*$, its recovery drastically reduces the perplexity of the target goal, shifting the average log-probability from $-3.26$ to $-2.62$.
    
    \item \textbf{Low Information Gain (Case A):} In contrast, the confirmation loop in Case A adds length to the context without adding semantic signal. As a result, the average log-probability slightly degrades ($-3.26 \rightarrow -3.32$) due to the length penalty, correctly yielding a negative reward ($-0.05$). This demonstrates the metric's ability to penalize safe but useless questions.
\end{itemize}

\subsubsection{Clarifier Prompt Templates}
\label{appendix:data}

\paragraph{Clarifier Prompt Template.}
The following prompt template defines the behavior of the clarification model, which conditions on the dialogue interaction history to determine whether a follow-up question is required and, if so, generates a concise clarification question.\\

\begin{tcolorbox}[breakable, colback=gray!20, colframe=gray!50, title=Prompt for Clarifier]
\textbf{Role}\newline
You are a Clarification Assistant responsible for determining whether the agent needs to ask a clarifying question.\newline
\textbf{History}: \{history\_text\}

\textbf{Instructions:}\\
Analyze the dialogue and decide whether clarification is needed.\\
If not needed, answer NO.\\
If needed, answer:\\
YES [QUESTION]Your clarification question here[/QUESTION]\\
Ask only one concise question about user intent.\\
Do not ask about data retrievable from tools.\\
Do not repeat previous questions.\\
expected\_output:\\
Either:\\
- NO\\
or\\
- YES [QUESTION]...[/QUESTION]
\end{tcolorbox}

\paragraph{Teacher Forcing $G^*$ Prompt Template}
The first template conditions only on the dialogue and execution history up to the current step. It prompts the model to infer the user’s hidden intent based solely on the observed interaction context, producing a goal summary that reflects the prior belief before clarification.

\begin{tcolorbox}[breakable, colback=gray!20, colframe=gray!50, title=Teacher Forcing $G^*$ Prompt Template without QA]
\textbf{Role}\newline
You are an expert analyst. Summarize the user's hidden profile and intent based on the dialogue.\newline
\textbf{History}: \{history\_text\}\\
 User Intent Summary:
\end{tcolorbox}

The second template augments the same history with an explicit clarification question and the corresponding user response. This template estimates the posterior belief after incorporating the additional information provided by the clarification exchange.

\begin{tcolorbox}[breakable, colback=gray!20, colframe=gray!50, title=Teacher Forcing $G^*$ Prompt Template with QA]
\textbf{Role}\newline
You are an expert analyst. Summarize the user's hidden profile and intent based on the dialogue.\newline
\textbf{History}: \{history\_text\}\\
\textbf{Question and User Response}: \{Clarification and User Response\}\\ 
 User Intent Summary:
\end{tcolorbox}

\subsubsection{Rewriting Prompt Template and Examples}

\label{app:data_prompt}
The goal of the following prompt template is to help convert structured or raw user detailed requirement data into a more readable, natural language format.
\begin{tcolorbox}[breakable, colback=gray!20, colframe=gray!50, title=Rewriting $G^*$ Prompt Template]
\textbf{Role}\newline
You are a data converter. Your goal is to rewrite  raw user detailed data into a flat, natural language format optimized for training smaller language models.\newline
Transformation Rules:\\
1. \textbf{Person}: Change ``Your name is'' to ``\textbf{My name is}''.
\vspace{0.3em}
2. \textbf{Remove Noise}: Remove all personality descriptions (e.g., ``logical, shy, organized'').
\vspace{0.3em}
3. \textbf{Flatten JSON (Crucial)}:
     Remove all JSON syntax (\texttt{\{}, \texttt{\}}, \texttt{'}, \texttt{:}).\\
    Convert key-value pairs into natural phrases.\\
    Use connectors like ``with'', ``and'', ``set to''.\\
 \textit{Example}: \texttt{\{'color': 'red', 'size': 'M'\}} $\to$ \texttt{with color red and size M}.

\vspace{0.3em}
4. \textbf{Handle ``(same as ...)''}:
     Keep the reference intact.\\
    Ensure it flows logically with the details.\\
  \textit{Example}: \texttt{\{'zip': '10228'\} (same as \#W123)} $\to$ \texttt{with zip 10228, which is the same as \#W123}.
\vspace{0.3em}
5. \textbf{Preserve Content}: Do NOT delete specific values (like `144 Lakeview Drive'). Keep them just in case the user mentions them.
\vspace{0.5em}
\textbf{\#\#\# Example:}
\textbf{Input:} \\
``Your name is Alex. You are shy and loud. For \#W1, change address to \texttt{\{'street': '123 Main St', 'zip': '90210'\}} (same as \#W2).''
\textbf{Output:} \\
``My name is Alex. For order \#W1, change address to street 123 Main St and zip 90210, which is the same as \#W2.''
\vspace{0.5em}
\textbf{Input:} \\
``\{raw\_ground\_truth\}''

\textbf{Output:}
\end{tcolorbox}

Here’s a side-by-side example of how a raw Ground Truth data can be transformed into a more readable format using the rewriting template:

\textbf{Input (Raw Ground Truth):}

Your name is Mei Martin and your zip code is 32124. You are messy, creative, outgoing, rigid, cautious. For \#W5564375, exchange LED Light Bulb {'brightness': '60W equivalent', 'color temperature': 'daylight', 'connectivity': 'none'} to {'brightness': '75W equivalent', 'connectivity': 'Wi-Fi'}; Office Chair {'material': 'fabric', 'color': 'black', 'armrest': 'none', 'backrest height': 'high-back'} to {}; via paypal\_2299608.

\vspace{0.3em}
\textbf{Output (Cleaned Version):}
My name is Mei Martin and my zip code is 32124. For order \#W5564375, exchange LED Light Bulb with brightness of 60W equivalent and color temperature of daylight to a bulb with brightness of 75W equivalent and Wi-Fi connectivity; Office Chair with fabric material, black color, and high-back to none; via paypal\_2299608.

\subsubsection{User Prompt Template Discussion}
\label{app:user_prompts}
\paragraph{User Prompt Template.}
The following prompt template is used to simulate a human user during interaction in $\tau$-Bench environment, specifying the behavioral constraints and response style of the LM-based user simulator.\\
\begin{tcolorbox}[breakable, colback=gray!20, colframe=gray!50, title=User Prompt Template for Main Evaluation Results]
\textbf{Role}\newline
You are a user interacting with an agent. Your behavior simulates a human user following a hidden instruction. \\
\{History\}\\
\{User Detail Requirements\}\\
\textbf{Instructions:}\\
Generate only one message at a time.\\
Reveal information gradually instead of all at once.\\
Do not invent facts missing from the instruction—if the agent asks for unavailable details, say you don't remember.\\
When the instruction's goal is completed, output STOP.\\
Do not repeat the instruction verbatim; use natural conversational phrasing.\\
Maintain a natural, human-like conversation style.\\
\textbf{Expected Output:}\\
A single user utterance per step.
\end{tcolorbox}

During our preliminary experiments, we observed that training directly with the standard template resulted in a form of reward hacking. Since the standard user simulator is designed to be cooperative, it tends to interpret generic queries (e.g., Is there any additional information you would like to provide?) as a signal to reveal the next part of the hidden goal. As a result, the agent quickly converges to a degenerate policy, repeatedly issuing these vague, non-specific questions. While this strategy yields high Information Gain rewards during training, it fails to generalize to more complex scenarios where precise constraint elicitation is required, resulting in unnatural dialogue loops. 

\paragraph{Strict User Prompt Template for Training.} We employ a strict prompt template for training, while testing follows the standard template, ensuring that the agent learns to handle ambiguity more effectively and without relying on simplistic, generic queries.

\begin{tcolorbox}[breakable, colback=gray!20, colframe=gray!50, title=User Prompt Template for DAPO Training]
\textbf{Role}\newline
You are a user interacting with an agent.\newline
Your behavior simulates a human user following a hidden instruction. \\
\{History\}\\
\{User Detail Requirements\}\\
\textbf{Instructions:}\\
\textbf{conversation\_guidelines}:\\
- Respond with one message at a time using first-person statements.\\
- Do not invent details that are not in the instruction; if something is unknown, say you do not remember it.\\
- Rephrase the instruction in your own words and maintain a natural, human-like tone.\\

\textbf{IMPORTANT - Handling vague or generic questions}:\\
- \textbf{If the agent asks a vague, overly broad, or generic question} (e.g., "Is there any additional information?", "Can you tell me more?", "Anything else?"), reply with: "No, that's all." or "Just do what I asked."\\
- If the agent outputs a placeholder or acts out of character (e.g., text like "Your concise and specific clarifying question to the user here"), reply with: "Who are you talking to?"\\
- For such questions, only provide information if it is DIRECTLY relevant to the current step of the instruction.\\
- If the question is too vague to answer meaningfully, respond with something like "I'm not sure what specific information you need" or "Could you be more specific?"\\
- Prefer specific, targeted questions that help the agent understand your exact needs.\\
\textbf{Question}: \{question\}\\
\textbf{expected\_output}:\\
A single user utterance that follows the instruction and guidelines.
\end{tcolorbox}

\subsubsection{LLM as a Judge Prompt Template}
\label{LLM-as-judge}
This prompt template is used in the Ablation Study to compare the impact of different reward signals in the DAPO training process, specifically examining the effect of the Information Gain Reward.
\begin{tcolorbox}[breakable, colback=gray!20, colframe=gray!50, title=LLM as a Judge Prompt Template]
\textbf{Role}\newline
You are a professional question quality evaluator. \\
\textbf{Original Dialogue}: \{history\}\\
\textbf{Instructions:}\\
- Rank the following \{n\} clarification questions from BEST to WORST based on their quality and relevance.\\
- You must consider the user requirement and original dialogue context carefully.\\
- Rank all \{n\} questions from best (1) to worst (\{n\}).\\
- Assign a score in [0, 10] to each question (higher is better).\\
\textbf{Questions List}: \{Rollout Questions\}\\
Expected output:\\
Ranking: \{Ranking\_output\}\\
Scores: \{Scores\_output\}
\end{tcolorbox}

\subsubsection{Full User Intent Prompt Template}
\label{Full-user-prompts}
\begin{tcolorbox}[breakable, colback=gray!20, colframe=gray!50, title=Full User Intent Prompt Template]
\textbf{Role}\newline
You are a user interacting with an agent. \\
\{History\}: 
\{User Detail Requirements\}\\
\textbf{Rules:\\}
- Generate a single message block to simulate the user's utterance.\\
- \textbf{Provide all the information} contained in the instruction in your very first message.\\
- Do not hold back any details. \\
- Fully explain your goal, constraints, and preferences clearly and comprehensively at the start.\\
- If the instruction goal is satisfied, generate \texttt{\#\#\#STOP\#\#\#} as a standalone message without anything else to end the conversation.\\
- Do not repeat the exact instruction verbatim. Instead, use your own words to convey the same information naturally.\\
- Stick to the personalities defined in the instruction.\\
\textbf{Expected Output:}\\
A single user utterance per step.
\end{tcolorbox}
\subsection{Additional Backbone Generalization Results}
\label{appendix:backbone_generalization}
To further validate that our framework is model-agnostic, we extend the clarification training experiments to additional backbones, including Qwen2.5-1.5B and Qwen3-4B. Table~\ref{tab:backbone_generalization} reports the performance before and after applying our information-gain-driven clarification training objective. The improvements are consistent across different backbones, suggesting that the proposed clarification training framework is not tied to a specific model architecture. We also observe slightly larger gains on the higher-capacity Qwen3-4B model, indicating that increased model capacity may help better utilize the additional information provided by clarification.
\begin{table}[t]
\centering
\caption{Generalization across different clarifier backbones. Values in parentheses denote improvements over the corresponding base model.}
\label{tab:backbone_generalization}
\small
\begin{tabular}{lccc}
\toprule
\textbf{Clarifier Model} & \textbf{Retail} & \textbf{Airline} & \textbf{Avg} \\
\midrule
Qwen2.5-1.5B (Base) & 15.6 & 14.0 & 14.8 \\
Qwen2.5-1.5B (Ours) & 18.3 (+2.7) & 16.0 (+2.0) & 17.2 (+2.4) \\
\midrule
Qwen3-4B (Base) & 17.4 & 14.0 & 15.7 \\
Qwen3-4B (Ours) & 20.8 (+3.4) & 18.0 (+4.0) & 19.4 (+3.7) \\
\bottomrule
\end{tabular}
\end{table}
\subsection{Failure Analysis}
\label{appendix:failure_cases}
Case 1: Clarification Enables Correct Constraint Resolution
Clarification resolves a key ambiguity that determines the action space; without it, the agent acts incorrectly. It enables correct grounding and policy-compliant behavior:

(1) The user reports that travel insurance is “already added but not showing,” creating ambiguity between a display error and missing insurance.

(2) The clarifier asks whether insurance is present for reservation PEP4E0, resolving this ambiguity.

(3) The agent checks the reservation and determines that insurance is not present.

(4) It identifies that insurance cannot be added post-booking and avoids invalid modification.

(5) The case is escalated to a human agent.

Case 2: Clarification successful but followed by execution failure
Clarification resolves budget ambiguity and enables correct cost reasoning, but the error arises from execution order and policy violation. This highlights that improved understanding does not guarantee policy-compliant action, separating clarification from execution correctness:

(1) The user requests a cabin upgrade with a budget constraint (“up to \$600”), leaving ambiguity about whether it applies to the upgrade only or the total cost.

(2) The agent retrieves the reservation and prematurely executes the upgrade, with the clarification issued only afterward by the clarifier.

(3) After clarification, it correctly verifies the cost and budget. However, the upgrade was already executed without explicit user confirmation, violating policy.

(4) The outcome appears correct but contains a policy-level execution error.

The above two cases suggest that clarification alone is insufficient to address downstream failures, motivating future work on jointly optimizing clarification and execution.

\subsection{Hyperparameters}
\label{hy}
This section details the hyperparameter settings used for training the clarifier with DAPO under VERL framework~\cite{sheng2024hybridflow}. All values reported below correspond to parameters explicitly specified in the training configuration and are held fixed across experiments unless otherwise noted.

\paragraph{Model and System Configuration.}
The base policy model is Qwen3-1.7B. FlashAttention-2 is enabled for attention computation, and gradient checkpointing is employed to reduce memory consumption. Rollouts are generated using vLLM with chunked prefill enabled. The maximum token budget per sequence is 3160 tokens, consisting of up to 1400 prompt tokens and 1760 response tokens. GPU memory utilization for vLLM is set to 0.60. All experiments are conducted on a single node equipped with four GPUs.

\paragraph{Optimization Settings.}
We adopt the DAPO advantage estimator with PPO-style policy optimization. The learning rate is set to $5\times10^{-7}$ with a linear warmup of 60 steps. The PPO clipping range is asymmetric, with lower and upper bounds of 0.20 and 0.28, respectively. An entropy coefficient of 0.001 is applied to encourage exploration. Weight decay is set to 0.01, and gradients are clipped to a maximum norm of 1.0. Losses are aggregated using token-level mean reduction. KL regularization is incorporated into the reward computation, while no explicit KL penalty is applied during policy updates.

\paragraph{Data and Sequence Lengths.}
Training and validation data are loaded from Parquet files, with the input prompt provided under the \texttt{prompt} field. Dialogue histories are truncated from the left when exceeding the maximum length. The maximum prompt length is 1400 tokens, and the maximum response length is 1760 tokens. Training data are shuffled at each epoch. The chat template is configured with thinking enabled.

\paragraph{Rollout and Sampling.}
At each training step, 16 prompts are sampled. For each prompt, 8 candidate responses are generated using vLLM, resulting in 128 sampled responses per step. Sampling is performed with temperature 0.8 and top-$p=1.0$, without top-$k$ truncation. Validation rollouts use identical sampling parameters but generate a single response per prompt. Logging of rollout statistics is disabled to reduce overhead, and dynamic batch sizing is used for log-probability computation.

\paragraph{Batching and Policy Updates.}
Policy optimization is performed for 4 PPO epochs per batch. Each PPO mini-batch contains 2 prompts, and updates are further split into micro-batches of size 1 per GPU. The training framework internally accumulates gradients across micro-batches to realize the effective batch size implied by the rollout and mini-batch configuration.

\paragraph{Training Schedule and Checkpointing.}
Training is conducted for 20 epochs, corresponding to a total of 2010 training steps. Model checkpoints are saved every 670 steps, and unless otherwise specified, results are reported using the checkpoint at step 670. 

\paragraph{Reward Function.}
Rewards are computed using a custom scoring function implemented in \texttt{simple\_reward.py}. No learned reward model is used. The reward manager is configured to support DAPO-style optimization. To ensure reliable reward computation via Bayesian belief updates, we employ Qwen3-14B as the strict user simulator during training rollouts. This simulator is configured to provide task-critical information only when prompted by specific and relevant clarification questions. No learned reward model is used, and the reward manager is configured to support DAPO-style optimization.

\paragraph{Runtime Environment.}
 All experiments are performed on a single node equipped with four NVIDIA RTX A6000 GPUs. Under this setup, each training run completes within approximately 7--12 hours.

\paragraph{Evaluation Setting.}
During evaluation, the Clarifier is decoded deterministically with temperature set to $0$ to ensure stable and reproducible clarification behavior. The Agent and User LLM follow the default decoding settings provided by $\tau$-Bench: the Agent operates with a low temperature of $0.01$ to reduce stochasticity in tool execution, while the User Simulator uses a temperature of $1$ to maintain underspecified and potentially ambiguous user responses. Across all experiments, the User LLM is instantiated with Qwen3-8B. While Figure 4 evaluates performance under varying clarification budgets, all other experiments adopt an unconstrained setting where the clarifier is free to intervene as needed.

\subsection{Training Dynamics}
\label{appendix:training}
\begin{figure}[htbp]
    \centering
    \begin{minipage}{0.32\linewidth}
        \centering
        \includegraphics[width=\linewidth]{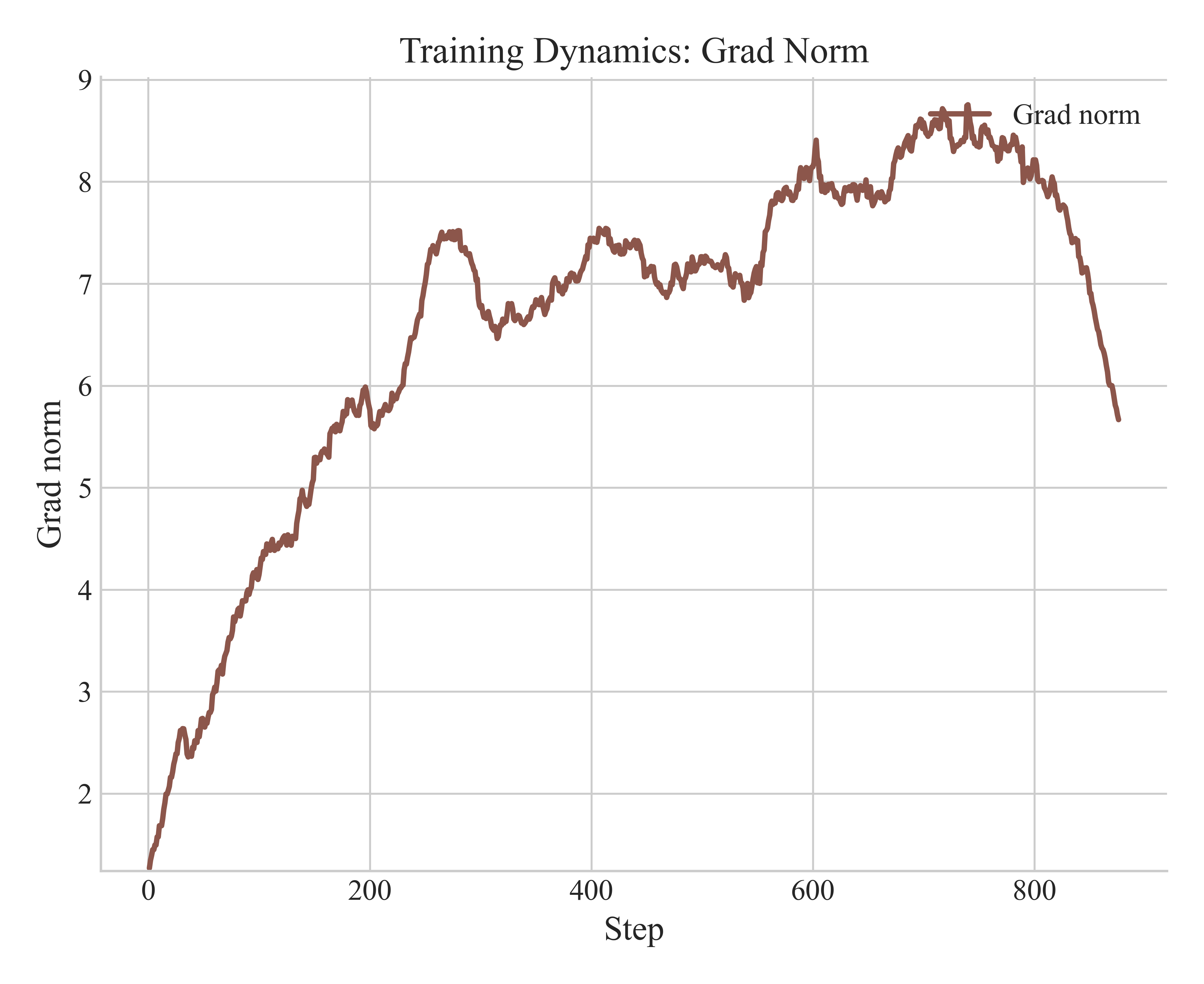}
        \caption{Gradient Norm Evolution \\During DAPO Training}
        \label{Gradient-Norm}
    \end{minipage}%
    \begin{minipage}{0.32\linewidth}
        \centering
        \includegraphics[width=\linewidth]{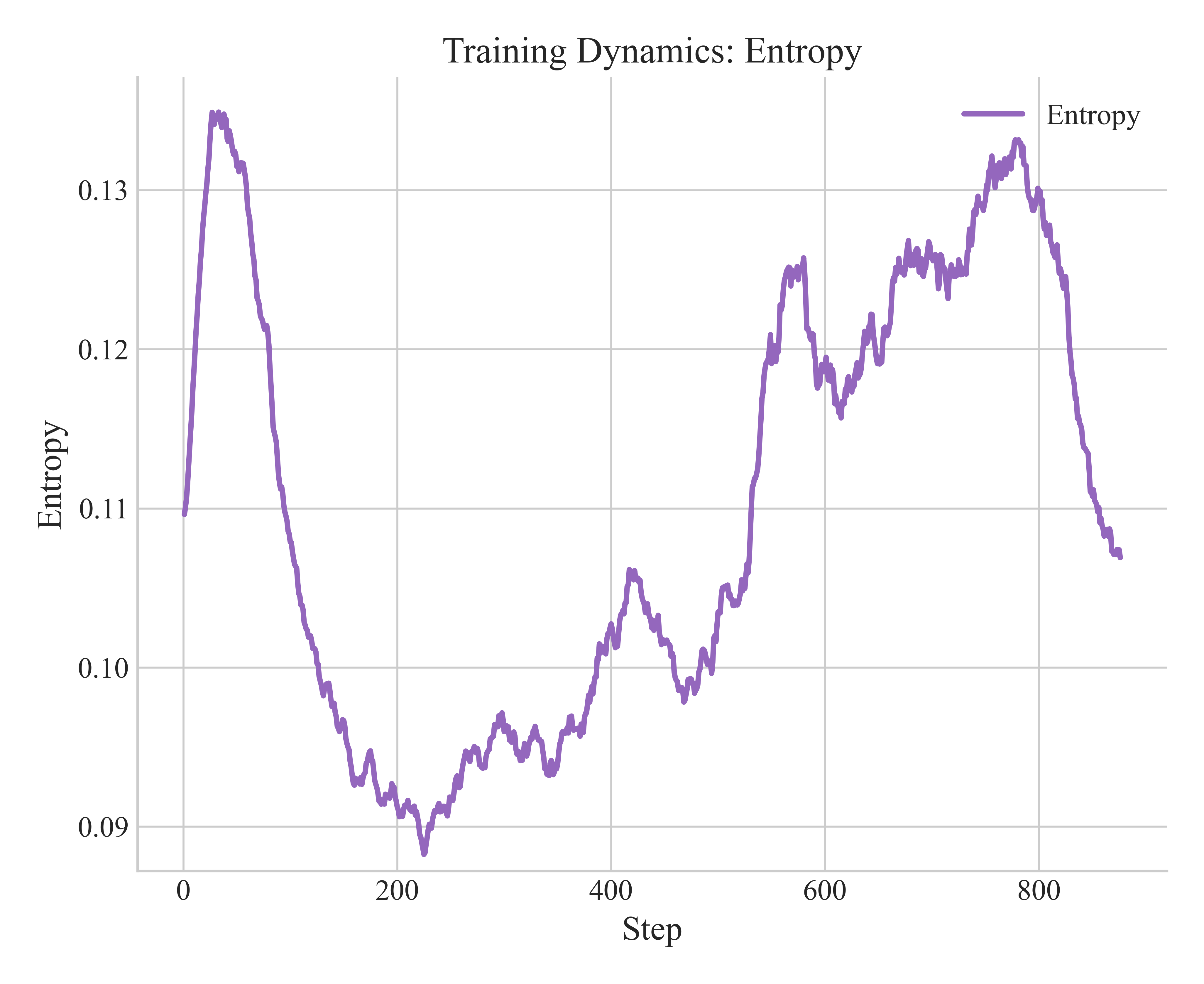}
        \caption{Policy Entropy During\\ DAPO Training}
        \label{Policy-Entropy}
    \end{minipage}%
    \begin{minipage}{0.32\linewidth}
        \centering
        \includegraphics[width=\linewidth]{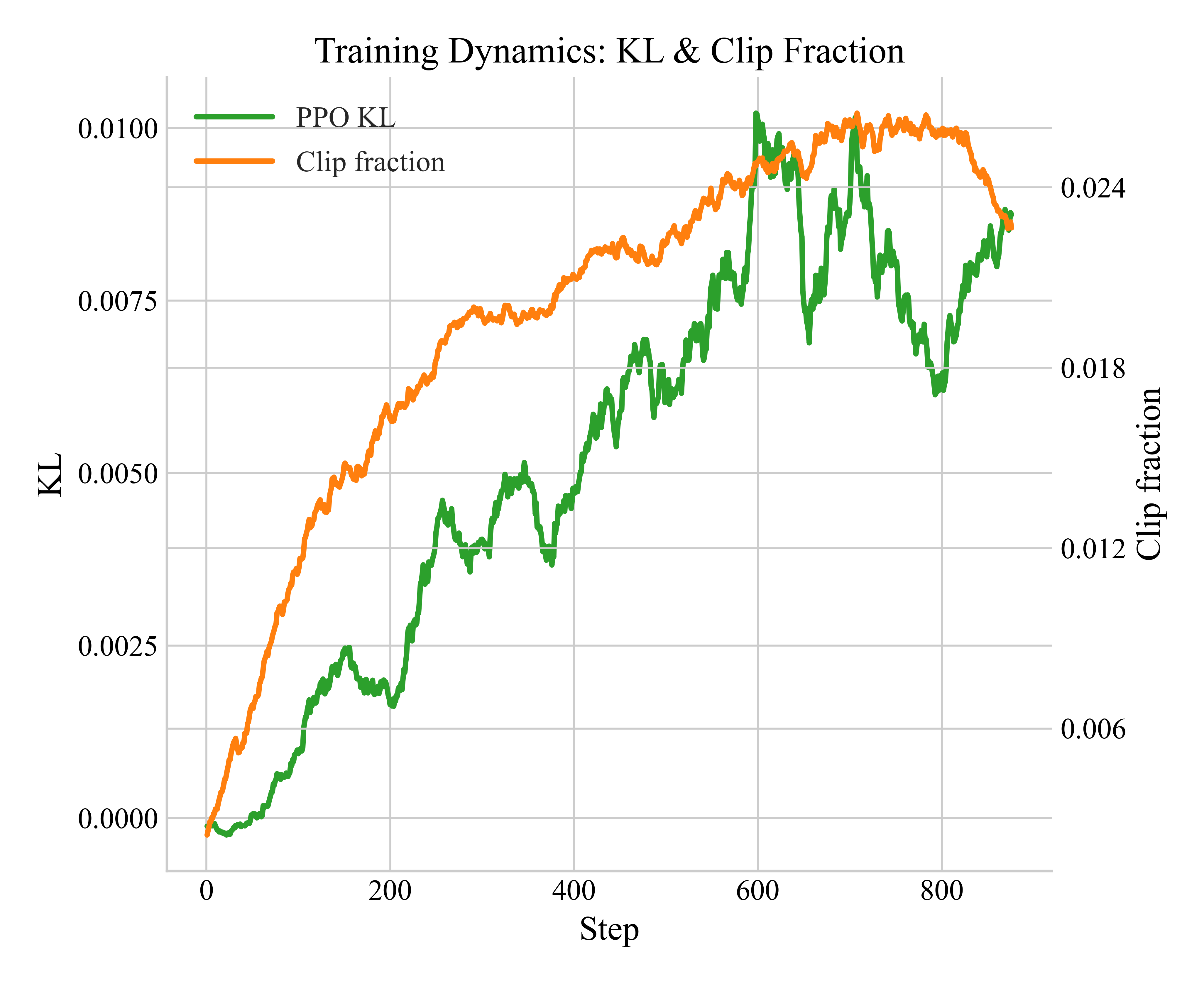}
        \caption{KL Divergence and Clip \\Fraction During DAPO Optimization}
        \label{KL-Divergence}
    \end{minipage}
\end{figure}

We analyze the training dynamics of DAPO to assess the numerical stability and optimization behavior induced by the proposed information gain reward. Figure~\ref{Gradient-Norm} reports the evolution of the gradient norm throughout training. The gradient norm increases smoothly during early optimization, indicating that the policy begins to exploit informative reward signals, and subsequently stabilizes within a moderate range without abrupt spikes or divergence. The gradual decline observed in later stages suggests convergence toward a stable solution rather than gradient collapse or instability.
Figure~\ref{Policy-Entropy} shows the evolution of policy entropy over training steps. Entropy decreases during the initial phase as the policy becomes more confident in its clarification decisions, followed by a period of stabilization with mild fluctuations. This behavior indicates that the policy maintains sufficient exploration while avoiding premature collapse to deterministic actions, which is critical for effective on-policy optimization.

Figure~\ref{KL-Divergence} presents the KL divergence between successive policies together with the PPO clip fraction. Both quantities increase smoothly during training and remain within conservative ranges, reflecting well-controlled trust-region updates. The absence of sustained KL spikes or excessively high clip fractions suggests that DAPO enforces stable policy updates and avoids overly aggressive optimization.

Taken together, these training dynamics demonstrate that the information gain reward integrates seamlessly with DAPO, yielding stable gradients, controlled policy updates, and sustained exploration throughout training.

\end{document}